\documentclass{article}

\usepackage{geometry}
\usepackage[utf8]{inputenc} 
\usepackage[T1]{fontenc}    
\usepackage{url}            
\usepackage{booktabs}       
\usepackage{amsfonts}       
\usepackage{nicefrac}       
\usepackage{microtype}      
\usepackage{natbib}

\usepackage[dvipsnames]{xcolor}
\usepackage[algo2e,ruled,vlined]{algorithm2e}

\SetCommentSty{mycommfont}
\usepackage{setspace}

\usepackage{amsmath}
\usepackage{wasysym}
\usepackage{graphicx}
\usepackage{algorithm}
\usepackage{algorithmic}
\usepackage{lipsum}
\renewcommand{\citet}{\citep}

\renewcommand{\cite}[1]{\citep{#1}}

\newcommand{\numdefenses}{thirteen }

\newcommand{\howitworks}{\subsection{How The Defense Works}}
\newcommand{\whyinteresting}{\paragraph{Why we selected this defense.}}
\newcommand{\hypothesis}{\subsection{Initial Hypotheses and Experiments}}
\newcommand{\attack}{\subsection{Final Robustness Evaluation}}
\newcommand{\exercise}{\paragraph{Exercises for the reader.}}
\newcommand{\lessons}{\paragraph{Lessons Learned.}}

\newcommand{\logits}{z}
\DeclareMathOperator*{\argmax}{arg\,max}

\usepackage{enumitem}
\newenvironment{myitemize}
{\begin{itemize}[leftmargin=5.5mm, itemsep=2pt, topsep=5pt]}
	{\end{itemize}}

\newenvironment{myenumerate}
{\begin{enumerate}[leftmargin=5.5mm, itemsep=1pt, topsep=5pt]}
	{\end{enumerate}}


\usepackage[hidelinks]{hyperref}       
\usepackage[capitalize]{cleveref}

\begin{document}

\title{On Adaptive Attacks \\ to Adversarial Example Defenses}
\author{Florian Tram\`er\thanks{Equal contribution}\\ Stanford University \and {Nicholas Carlini}$^*$ \\ Google \and {Wieland Brendel}$^*$ \\ University of T\"ubingen \and {Aleksander M\k{a}dry} \\ MIT}

\date{}
\maketitle

\begin{abstract}
  Adaptive attacks have (rightfully) become the de facto standard for evaluating
  defenses to adversarial examples.
  We find, however, that typical adaptive evaluations
  are incomplete.
  We demonstrate that \numdefenses defenses recently published at ICLR, ICML and NeurIPS---and which illustrate a diverse set of defense strategies---can be circumvented {\em despite} attempting to perform evaluations using adaptive attacks.
  While prior evaluation papers focused mainly on the end
  result---showing that a defense was ineffective---this
  paper focuses on laying out the methodology and the approach
  necessary to perform an adaptive attack.
  Some of our attack strategies are generalizable, but no single strategy would have been sufficient for all defenses. This underlines our key message that adaptive attacks {\em cannot} be automated and always require careful and appropriate tuning to a given defense.
  We hope that these analyses will serve as guidance on how
  to properly perform adaptive attacks against defenses to adversarial
  examples, and thus will allow the community
  to make further progress in building more robust models.
\end{abstract}

\newpage
\setcounter{tocdepth}{1}
\tableofcontents
\newpage

\section{Introduction}
Over the last five years the research community has attempted to develop
defenses to adversarial examples \citep{szegedy2013intriguing,biggio2013evasion}.
This has proven extraordinarily difficult.
Indeed, a common theme has been proposing defenses that---due to having been tested only against static and relatively weak
attacks---were promptly circumvented by a stronger attack~\citep{carlini2017adversarial,athalye2018obfuscated}.

Recent community efforts and guidelines to improve defense evaluations have had a noticeably positive impact.
In particular, there has been a significant uptake of evaluations against \emph{adaptive attacks}, i.e., attacks that were specifically designed to target a given defense---the ratio of defenses evaluated
against adaptive attacks has increased from close to zero in 2017 \citep{carlini2017adversarial}
to one third in 2018 \citep{athalye2018obfuscated} and to nearly all of them today.%
\footnote{There is a similarly positive trend in terms of releasing source code for defenses.  In particular, {\em every} defense we analyzed either released
source code, or made it available upon request.}
This leads to the question:

\begin{center}
\emph{With their much-improved evaluation practices, are these defenses truly robust?}
\end{center}

\noindent We find that this is {\em not} the case.
Specifically, in an analysis of \numdefenses defenses, 
selected from recent ICLR, ICML, and NeurIPS conferences to illustrate diverse defensive strategies,
we find that we can circumvent \emph{all} of them and substantially
reduce the accuracy from what was originally claimed.

Importantly, while almost all of these defenses performed an evaluation involving adaptive attacks, these evaluations ended up not being sufficient.
For example, it is common for papers to repurpose existing ``adaptive'' attacks that circumvented some prior defense, without considering how to change it to target the new defense.
We suspect that this shortcoming might have been caused, in part, by the fact that prior work on circumventing defenses typically shows only the final, successful attack, without describing the methodology that was used to come up with this attack and thus leaving open questions such as ``How was the attack discovered?'' or ``What other attacks were unsuccessful?''. 

To remedy this problem, instead of merely demonstrating that the \numdefenses defenses we studied can be circumvented by stronger attacks,
we actually walk the reader through our full process of analyzing each defense, from an initial paper read-through, to our hypotheses about what would be required for the defense to be circumvented, to an ultimately successful attack. This approach lets us more clearly document the many steps involved in
developing a strong adaptive attack.

The goal of our analyzes is not to reduce analyzed model's
accuracy all the way to $0\%$.%
\footnote{That said, most of the defenses we study are ineffective
at increasing robustness even partially, and we are able to reduce
the accuracy of the defended classifier to the accuracy of a baseline model.}
Instead, we want to demonstrate that the existing adaptive attack evaluation methodology has shortcomings, and that stronger adaptive attacks can (at least partially) degrade each defense's accuracy from what is reported in the original evaluations.%

Whereas prior work often needed to develop new techniques \cite{athalye2018obfuscated}
to evade defenses, we find that the
technical tools to evaluate defenses properly already exist.
A better attack can be built using only tools that are well-known in the literature. 
Thus, the issue with current defense evaluations is methodological rather than technical.

After describing our methodology (\cref{sec:methodology}) and providing an overview of common themes distilled from our evaluations (\cref{sec:overview}), each subsequent section of this paper is dedicated exclusively to the full evaluation of one defense.
We state and test
our initial hypotheses---gathered from reading the paper and source code---as to why the original
evaluation may have been insufficient.
Finally, we describe how the observations made 
throughout our evaluation inform the design of a final adaptive attack that
succeeds in circumventing the defense.

An overarching theme of our evaluations is {simplicity}.
Attacks need not be complicated, even when the defense is.
There are often only a small number of important components in complex defenses---carefully targeting these can lead to simpler and stronger attacks.
We design our loss functions, the cornerstone of successful adaptive attacks,
so that they are easy to optimize and \emph{consistent}---so that higher loss values result in strictly stronger attacks.
While some of our techniques are generalizable, no single attack strategy would have sufficed for all defenses. This underlines the crucial fact that adaptive attacks {\em cannot} be automated and always require appropriate tuning to a each defense.

In the spirit of responsible disclosure, we contacted all the authors of the defenses we evaluated prior to this submission, and offered to share code or adversarial examples to enable independent verification of our claims. In all but one case (where the defense authors did not respond to our original message), the authors acknowledged our attacks' effectiveness. To promote reproducibility and encourage others to perform independent re-evaluations of proposed defenses, we release code for all of our attacks at \url{https://github.com/wielandbrendel/adaptive_attacks_paper}.

\section{Background}
\label{sec:background}

This paper studies the robustness of defenses to adversarial examples.
Readers familiar with the relevant literature and notation \citep{szegedy2013intriguing,carlini2017towards,madry2017towards,athalye2018obfuscated}
can continue with Section~\ref{sec:methodology} where we describe our
methodology.

For a classification neural network $f$ and a natural
input $x$ (e.g., drawn from the test set)
with a true label $y$, an \emph{adversarial example}
\citep{szegedy2013intriguing} is a perturbed input
$x'$ such that:
(1) $\lVert x' - x \rVert$ is small, for
some distance function\footnote{We evaluate each defense under the threat model for which robustness
  was claimed; in all cases $\ell_p$ norms for various $p\ge0$.}
$\lVert \cdot \rVert$
but (2) it is classified incorrectly either \emph{untargeted}, so that $f(x') \ne y$,
or \emph{targeted}, so that $f(x') = t$ for some selected class $t \ne y$.\footnote{%
Most papers evaluate on whichever result is stronger:
thus, papers that propose attacks perform targeted attacks
(because if a targeted attack succeeds, then so will untargeted attacks)
and papers that propose defenses perform untargeted attacks
(because a model robust to untargeted attacks is also
robust to targeted attacks).
We intend for this paper to be a study in 
how to perform adaptive attacks evaluations; as such, we choose the untargeted
attack objective that future defense creators will typically choose.
}

To generate an adversarial example, we construct a \emph{loss function} $L$ so that
$L(x, y)$ is large when $f(x) \ne y$, and then 
maximize $L(x', y)$ while keeping the perturbation
$\lVert x - x' \rVert$ small.
Defining an appropriate loss function $L$ is the key component of any
adaptive attack.
The typical starting-point for any loss function is the
cross-entropy loss $L_{CE}$.

While there are many methods for generating adversarial examples,
the most widely adopted approach is gradient descent on
input space \citep{carlini2017towards,madry2017towards}.
Let $x_0 = x$ and then repeatedly set
\[x_{i+1} = \texttt{Proj}(x_i + \alpha \cdot \texttt{normalize}(\nabla_{x_i} L(x_i, y)))\]
where $\texttt{Proj}$ projects the inputs onto a smaller domain
to keep the distortion small,
$\texttt{normalize}$ enforces a unit-length step size under the considered norm,
and $\alpha$ controls the size of steps that are taken.
After $N$ steps (e.g., 1000), we let $x'=x_N$ be the resulting adversarial example. 
Further background on common adversarial examples generation techniques is provided in \cref{sec:attack_background}.

In some cases, it will be useful to represent the classifier in the
form $f(x) = \argmax \logits(x)$, where $\logits(x)$ is a vector
of class-scores which we refer to as the logit layer
(e.g., before any softmax activation function); thus, $\logits(x)_i$
is the logit value for class $i$ on input $x$. 
Through a slight abuse of notation, we sometimes use $f(x)$ to refer to 
either the classifier's output class (i.e., $f(x)=y$) or to the full vector of class probabilities (i.e., $f(x)=[p_1, \dots, p_K]$). The choice should be clear from context.

All defenses we study in this paper claim \emph{white-box} robustness:
here, the adversary is assumed to have knowledge of the model architecture,
weights, and any necessary additional information.
This allows us to use the above attack techniques.

\subsection{Background on Procedures for Generating Adversarial Examples}
\label{sec:attack_background}

This section provides some additional background on common attack strategies used in defense evaluations.

\paragraph{PGD.}

Projected Gradient Descent~\citep{madry2017towards} is a strategy for finding an adversarial example $x'$ for an input $x$ that satisfies a given norm-bound $\|x' -x\|_p \leq \epsilon$.

Let $B$ denote the $\ell_p$-ball of radius $\epsilon$ centered at $x$.
The attack starts at a random point $x_0 \in B$, and repeatedly sets
\begin{align*}
x_{i+1} &= \texttt{Proj}_{B}(x_i + \alpha \cdot g) \\
\quad \text{for } g &= \argmax_{\|v\|_p \leq 1} v^\top \nabla_{x_i} L(x_i, y) \;.
\end{align*}
Here, $L(x, y)$ is a suitable loss-function (e.g., cross-entropy), $\alpha$ is a step-size, $\texttt{Proj}_B$ projects an input onto the norm-ball $B$, and $g$ is the \emph{steepest ascent} direction for a given $\ell_p$-norm. E.g., for the $\ell_\infty$-norm, $\texttt{Proj}(z)$ is a clipping operator and $g = \texttt{sign}(\nabla_{x_i} L(x_i, y))$ .

\paragraph{C\&W.}

\citet{carlini2017towards} propose a number of attacks commonly referred to as C\&W attacks. Instead of maximizing a loss function $L(x, y)$ under a given perturbation constraint (as is done with PGD above), these attacks aim to find the \emph{smallest} successful adversarial perturbation. That is, the attack maximizes the following objective:
\[
\text{maximize}_{x'} \ L(x', y) - \lambda \cdot \|x'-x\|_p \;.
\]
The parameter $\lambda>0$ balances the objectives of maximizing the loss and minimizing the perturbation size. A binary-search over $\lambda$ is used to find the minimally perturbed $x'$ that results in a successful attack.

\paragraph{BPDA.}

The Backward Pass Differentiable Approximation (BPDA) is a gradient approximation technique introduced by~\citet{athalye2018obfuscated}. It is suited for defenses that have one or more non-differentiable components.

Let $f$ be a $n$-layer network $f(x) = f^n \circ f^{n-1} \circ \dots \circ f^1(x)$ where layer $f^i(\cdot)$ is non-differentiable (or hard to differentiate).
To approximate the gradient $\nabla_x f(x)$, we first find a differentiable function $g(x)$ such that $g(x) \approx f^i(x)$. Then, when computing the gradient $\nabla_x f(x)$ via backpropagation, we compute a standard forward pass through $f(\cdot)$ (including the forward pass through $f^i(\cdot)$), but replace $f^i(\cdot)$ by $g(\cdot)$ on the backward pass.

The approximated gradient can then be plugged into a standard gradient-based attack.
For defenses that apply some type of non-differentiable ``de-noising'' layer $f^1(x)$ to the input, it is often effective to approximate $f^1(x)$ with the identity function $g(x)=x$.

\paragraph{EOT.}

Expectation over Transformation~\citep{athalye2017synthesizing} is a standard technique for computing gradients of models with randomized components.
The attack was originally proposed for situations where a
randomized transformation is applied to an input $x$ before being fed into a classifier.
The idea is more generally applicable for obtaining gradients of the expectation of any randomized function.

Given a randomized classifier $f_r(x)$ (where $r$ denotes the classifier's internal randomness), we can compute
\[
\nabla_x \mathbb{E}_r [f_r(x)] = \mathbb{E}_r [\nabla_x f_r(x)] \approx \frac1n \sum_{i=1}^n \nabla_x f_{r_i}(x) \;,
\]
where the $r_i$ are independent draws of the function's randomness.
As with BPDA, the approximated gradients can then be plugged into any standard attack.

\section{Methodology}
\label{sec:methodology}

The core of this paper documents the attacks we develop on the \numdefenses
defenses. 
Our structure for each evaluation section is as follows:

\begin{myitemize}
\item \textbf{How the defense works.}
  We begin with a brief description of the defense and introduce
  the necessary components.
  We encourage readers to pause after this section to reflect on 
  how one might attack the described defense before reading our method.%
  \footnote{To facilitate the reading of this paper, we use the standard notation 
  defined in~\cref{sec:background} to describe each defense and attack. Each section introduces additional defense-specific notation only if needed. Our notation thus often differs slightly from that used in each defense's original paper.}

\item \textbf{Why we selected this defense.}
  While we study many defenses in this paper, we cannot hope to evaluate
  \emph{all} defenses published in the last few years.
  We selected each defense with two criteria:
  (1) it had been accepted at ICLR, ICML, or NeurIPS; indicating
  that independent reviewers considered it interesting and sound, and
  (2) it clearly illustrates some concept; if two defenses
  were built on a same idea or we believed would fail in
  identical situations we selected only one.\footnote{We discovered \emph{a posteriori} that some of the selected defenses can fail in similar ways. This report contains our evaluations 
  for \emph{all} of our initially selected defenses. In particular, we did not cherry-pick the defense evaluations to present based on the results of our evaluation.}

\item \textbf{Initial hypotheses and experiments.}
  We fist read each paper with a focus on finding 
  potential reasons why the defense might still
  be vulnerable to adversarial examples, in spite of
  the robustness evaluation initially performed in the paper.
  To develop additional attack hypotheses, we
  also studied the defense's source code, which
  was available publicly or upon request for \emph{all} defenses.

\item \textbf{Final robustness evaluation.}
  Given our candidate hypotheses of why the defense could fail,
  we turn to developing our own adaptive attacks.
  In most cases, this consists of (1)
  constructing an improved loss function where gradient descent could succeed
  at generating adversarial examples,
  (2) choosing a method of minimizing that loss function,
  and (3) repeating these steps based on new insights gleaned from an attack attempt.

\item \textbf{Lessons learned.}
  Having circumvented the defense, we look back and ask what
  we have learned from this attack and how it may apply in the future.
  
\end{myitemize}

The flow of each section directly mirrors the steps we actually took
to evaluate the robustness of the defense, without retroactively
drawing conclusions on how we wish we had discovered the flaws.
We hope that this will allow our evaluation strategy to
serve as a case study for how to perform future
evaluations from start to finish, rather than just observing the end result.

We emphasize that our main goal is to assess whether each defense's initial robustness evaluation was appropriate, and---if we find it was not---to demonstrate how to build a stronger adaptive attack on that defense. 
Importantly, we do not wish to claim that the robustness techniques used by many of these defenses hold no merit whatsoever. It may well be that some instantiations of these techniques can improve model robustness. To convincingly make such a claim however, a defense has to be backed up by a strong adaptive evaluation---the focal point of our tutorial.

\section{Recurring Attack Themes}
\label{sec:overview}

We identify six themes (and one meta-theme) that are common to multiple
evaluations. We discuss each below.
\cref{tab:themes} gives an overview of the themes that relate to each of the defenses we studied.

\textbf{(meta-theme) T0: Strive for simplicity.}
Our adaptive attacks are consistently \emph{simpler} than the adaptive
attacks evaluated in each paper.
In all cases, our attack is as close as possible to straightforward
gradient descent with an appropriate loss function;
we add additional complexity only when
simpler attacks fail.
It is easier to diagnose failures of simpler attacks,
in order to iterate towards a stronger attack.
The themes below mainly capture various ways to make an attack 
simpler, yet stronger.

\begin{table}[t]
	\centering
	\setlength{\tabcolsep}{5pt}
	\def\arraystretch{1}
	\begin{tabular}{@{}l l c c c c c c@{}}
		&& \multicolumn{6}{c}{\textbf{Attack Themes}} \\
		\cmidrule{3-8}
		\multicolumn{2}{c}{\textbf{Defense}} &  \textbf{T1} & \textbf{T2} & \textbf{T3} & \textbf{T4} & \textbf{T5} & \textbf{T6}\\
		\toprule
		\cref{sec:kwta}& \emph{k-Winners Take All}
		\citep{xiao2019resisting} 
		& \CIRCLE & & & & \CIRCLE & \CIRCLE\\
		\cref{sec:odds}& \emph{The Odds are Odd}
		\citep{roth2019odds} 
		 & && \CIRCLE & \CIRCLE &\\
		\cref{sec:generative}& \emph{Generative Classifiers}
		\citep{li2018generative} 
		 & & \CIRCLE & \CIRCLE &&\\
		\cref{sec:fourier}& \emph{Sparse Fourier Transform}
		\citep{bafna2018thwarting} 
		 & \CIRCLE & \CIRCLE & & & \\
		\cref{sec:rethinking}& \emph{Rethinking Cross Entropy}
		\citep{pang2020rethinking} 
		& & & \CIRCLE & & \CIRCLE \\
		\cref{sec:ecc}& \emph{Error Correcting Codes}
		\citep{verma2019error} 
		& \CIRCLE & \CIRCLE & & & \\
		\cref{sec:ensemble}& \emph{Ensemble Diversity}
		\citep{pang2019improving} 
		& & & & & \CIRCLE \\
		\cref{sec:empir}& \emph{EMPIR}
		\citep{sen2020empir} &
		& \CIRCLE & & & \CIRCLE &  \\
		\cref{sec:temporal}& \emph{Temporal Dependency}
		\citep{yang2018characterizing} 
		& \CIRCLE & & \CIRCLE & \CIRCLE & \CIRCLE \\
		\cref{sec:mixup}& \emph{Mixup Inference}
		\citep{pang2020mixup} 
		& \CIRCLE & & & & \\
		 \cref{sec:menet}& \emph{ME-Net}
		\citep{yang2019me}
		& \CIRCLE & \CIRCLE && \CIRCLE & &\CIRCLE\\
		\cref{sec:asymmetrical}& \emph{Asymmetrical Adv.~Training}
		\citep{yin2020adversarial} 
		&&&\CIRCLE & \CIRCLE & & \CIRCLE\\
		\cref{sec:weakness}& \emph{Weakness into a Strength}
		\citep{hu2019new} 
		 & \CIRCLE & \CIRCLE & \CIRCLE & \CIRCLE & \\
		 \bottomrule
	\end{tabular}
		\vspace{0.25em}
        \caption{The attack themes illustrated by each defense we evaluate.}
\label{tab:themes}
\end{table}

\textbf{T1: Attack (a function close to) the full defense.}
If a defense is end-to-end differentiable, the entire defense should
be attacked.
Additional loss terms should be avoided unless they are necessary;
conversely, any components, especially pre-processing functions, should be
included if at all possible.

\textbf{T2: Identify and target important defense parts.}
Some defenses combine many sub-components in a complex fashion. Often, inspecting these components reveals that only one or two would be sufficient for the defense to fail. Targeting only these components can lead to attacks that are simpler, stronger, and easier to optimize.

\textbf{T3: Adapt the objective to simplify the attack.}
There are many objective functions that can be optimized to
generate adversarial examples.
Choosing the ``best'' one is not trivial but can vastly boost an attack's success rate.
For example, we find that targeted attacks (or ``multi-targeted attacks''~\citep{gowal2019alternative}) are sometimes easier to formulate than
untargeted attacks.
We also show that \emph{feature adversaries}~\citep{sabour2015adversarial} 
are an effective way to circumvent many defenses that \emph{detect} adversarial examples.
Here, we pick a natural input $x^*$ from another class than $x$ and generate
an adversarial example $x'$ that matches the feature representation of $x^*$ at some inner layer.

\textbf{T4: Ensure the loss function is consistent.}
Prior work has shown that many defense evaluations fail because of loss functions that are hard to optimize~\citep{carlini2017towards, athalye2018obfuscated}.
We find that many defense evaluations suffer from an orthogonal and deeper issue: the chosen loss function is not actually a good proxy for attack success. That is, even with an all-powerful optimizer that is guaranteed to find the loss' optimum, an attack may fail. Such loss functions are not \emph{consistent} and should be avoided.

\textbf{T5: Optimize the loss function with different methods.}
Given a useful loss function that is as simple as possible, one should choose the appropriate attack algorithm as well as the right hyper-parameters (e.g. sufficiently many iterations or repetitions). In particular, a white-box gradient-based attack is 
not always the most appropriate algorithm for optimizing the loss function.
In some cases, we find that applying a ``black-box'' score-based attack~\citep{chen2017zoo,ilyas2018black} or decision-based attack~\citep{brendel2017decision}
can optimize loss functions that are difficult to differentiate.

\textbf{T6: Use a strong adaptive attack for adversarial training.}
Many papers claim that combining their defense with adversarial training~\cite{szegedy2013intriguing, madry2017towards} yields a stronger defense than either technique on its own. Yet, we find that many such combined defenses result in \emph{lower} robustness than adversarial training alone. This reveals another failure mode of weak adaptive attacks: if the attack used for training fails to reliably find adversarial examples, the model will not resist stronger attacks.

\section{k-Winners Take All}
\label{sec:kwta}

This paper~\citep{xiao2019resisting} proposes an activation function that is intentionally designed to mask backpropagating gradients in order to
defend against gradient-based attacks.
\howitworks

This defense replaces the standard ReLU activation function in a deep neural network with a discontinuous
k-Winners-Take-All ($k$-WTA) function,
\begin{equation*}
	\phi_k({\bf y})_j = \begin{cases}
y_j, & y_j\in \left\{k \mbox{ largest elements of } {\bf y}\right\},\\
0, & \mbox{else.}\\
\end{cases}
\end{equation*}
The activation function is applied to the output of a whole layer. The parameter $k$ is typically not the same for all layers. Instead, for each layer $k$ is 
obtained by multiplying the dimensionality $N$ of the layer's output with a constant sparsity factor $\gamma\in [0, 1]$.

The result of this particular choice of activation function is that even small changes to the input drastically change the activation patterns in the network and thus lead to large jumps
in the predictions. This behavior is highlighted in the manuscript~\citep[Figure 5]{xiao2019resisting}. This chaotic
partitioning of the activation patterns destroys all useful gradient information and thus prevents gradient-based
attacks from finding minimal adversarial examples.

\whyinteresting
There are many defenses that unintentionally make gradient descent hard.
This paper takes that to the extreme and designs the defense with a ``$C^0$
discontinuous function that purposely invalidates the neural network
model's gradient'' \citep{xiao2019resisting}.
Because prior work has argued that this form of gradient masking is
a fundamental flaw \citep{athalye2018obfuscated},
the existence of such a defense would go against common
wisdom.

\hypothesis
The implicit assumption of the paper is that gradient-based attacks
are \emph{strictly} stronger than black-box attacks such as score-based or decision-based attacks~\citep{chen2017zoo, brendel2017decision}.
There are many classes of functions for which this is not true.
For example, consider
any undefended network and add a simple non-differentiable activation function
to the final logits that quantizes them extremely finely.
Then, na\"ive gradient-based attacks would fail because all
gradients would be zero or undefined.
However, estimating the gradient from finite
differences would still work.\footnote{%
Note that an adaptive attacker could just remove such a
  non-differentiable layer and then succeed.
  While such a defense may sound implausible,
  \cref{sec:ecc} evaluates exactly such
  a defense that unintentionally post-processes with the ``identity''
  function $g(y) = \log(\exp(y))$ which, while mathematically a no-op,
  causes numerically unstable gradients.
  Removing this post-processing function breaks the defense.}

In order to demonstrate that the defense is effective, the paper
tests against transfer-based attacks.
While it argues that ``generating adversarial examples from an independently trained copy of the target network'' is ``the strongest black-box attack'',
in practice transfer attacks are often weak.
The success of transfer attacks strongly depends
on how close the substitute model is to the target model.
Training with the same architecture is insufficient; networks trained from different initializations can be far apart in their internal representations. The perturbation size necessary for transfer attacks to succeed is thus often significantly larger than for white-box attacks~\citep{papernot2016transferability, tramer2017ensemble}.
One setting where transfer-based attacks can be strong is if the model $f$ can be transformed into a nearly identical model $f'$ for which it is easier to generate
adversarial examples (e.g., the BPDA attack of~\citet{athalye2018obfuscated} can be seen as an instance of this approach).
In contrast,
even the most restricted class of direct attacks, 
decision-based attacks (which only use the final label
decision of the model~\citep{brendel2017decision}), can perform as well as gradient-based
attacks if the attack is run until convergence.

While the omission of score- and decision-based attacks is the biggest concern, a few other issues make it difficult to assess the true robustness of this defense from the initial evaluation.
First, the lack of a distortion-accuracy plot makes it difficult to spot gradient-masking issues~\citep{carlini2019evaluating}. Second, some attacks (C\&W, DeepFool) are actually optimizing the $\ell_2$ metric and not the $\ell_\infty$ metric under which the accuracy is measured, meaning that the values for those attacks should be much higher than for proper $\ell_\infty$ attacks (but they are often close, within a few percentage points). Finally, the attacks in the paper are only applied with a small number of steps (20-40), which might have prevented attacks from converging, and without any repetitions (which can often reduce accuracy by 10\% or more).

For our experiments, we evaluate the authors' released defense on CIFAR-10
with $\ell_\infty$ bounded attacks of distortion $\epsilon = 8/255$.
We evaluate $k$-WTA at sparsity ratio $\gamma=0.1$ as the paper reports the strongest results with this setting.
We evaluate both a vanilla and an adversarially trained ResNet-18 for which
the paper claims $13.1\%$ and $50.0\%$ accuracy (respectively).
All our experiments rely on implementations in Foolbox~\citep{rauber2017foolbox}.

We begin our evaluation by running a standard PGD attack~\citep{madry2017towards}, with
400 steps instead of 40 as performed in the paper's original evaluation.
We find that over the progression of the attack, the cross-entropy loss exhibits high fluctuations. In fact, adversarial examples were seemingly found by chance and the attack did not reliably decrease the average loss.
This suggests that the extreme jagged loss surface of the $k$-WTA-model
could even \emph{decrease} the accuracy compared to a baseline model:
because of the high fluctuations in the logits, it is possible that
even in correct and high-confidence regions of the original model,
the fluctuations in the logits could introduce adversarial examples.

\begin{figure}
  \caption{Loss landscape of the k-Winners Take All defense. The plot shows the difference between the logit of the ground-truth class and the next highest logit along a random perturbation direction $x + \epsilon\operatorname{sign}(\delta)$. Note the x-axis scale: the largest perturbation changes the input by 0.1\%.}
  \label{fig:kwta:losslandscape}
  \centering
  \vspace{0.7em}
    \includegraphics[width=0.6\textwidth]{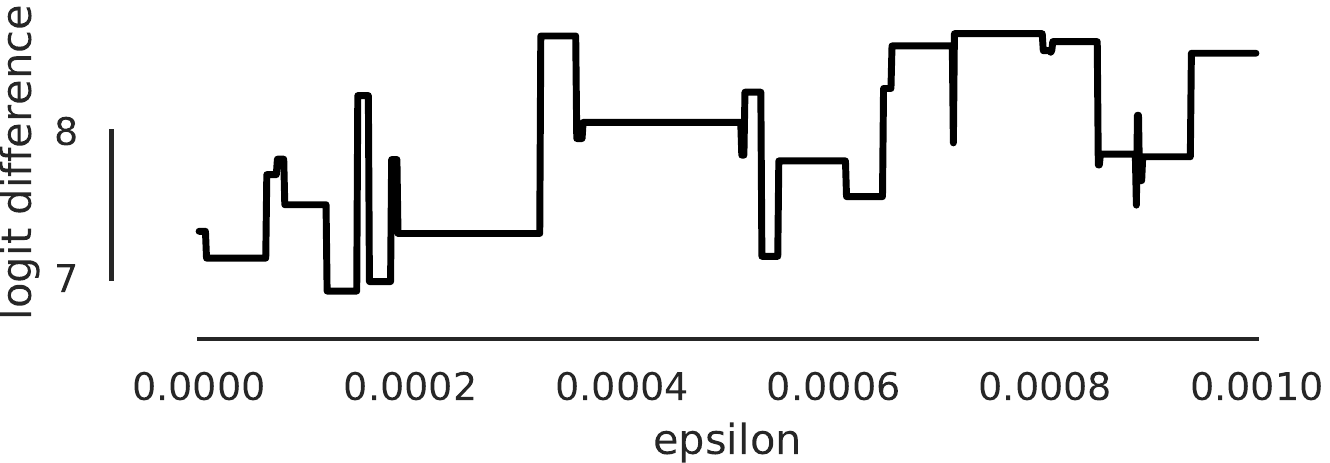}
\end{figure}

We then looked at the loss function close to clean samples. As is visible in \cref{fig:kwta:losslandscape}, even very small perturbations can flip the loss
to a new state. The loss stays roughly constant until it flips again if the perturbation
is further increased. This explains why standard gradient-based attacks fail:
the gradients can only highlight the direction of descent on each plateau but 
cannot take into account the substantial non-linear activation shifts at the boundary
of the plateaus. In other words, the gradients are only valid in an extremely small and
confined region but are not representative of the descent directions over larger regions.
\cref{sec:rethinking} contains another defense that fails for a similar reason.

We therefore hypothesize that a strong black-box attack that directly queries the target model and
estimates the gradients from finite differences over a larger space would be an effective attack on this defense.

\attack
Our final attack strategy was to average out the randomness by
estimating the average local gradient $\delta$ at point $x$ with
$M = 20000$ random perturbations drawn from a standard normal
distribution with standard deviation $\epsilon = 8/255$, i.e.,
\begin{equation*}
	g(x) = \frac{2}{\sigma M}\sum_{j=0}^{M/2} \left[\nabla_x L_{CE}(f(x + \delta), y) + \nabla_x L_{CE}(f(x - \delta), y)\right] \\
\end{equation*}
where $\delta\sim \mathcal{N}(\mu = 0, \sigma^2)$ and $L_{CE}$ is the cross-entropy loss.
We run this attack for 100 steps and successfully
generated adversarial examples for both
the vanilla as well as the adversarially trained ResNet-18 that
reach 0\% and 0.16\% accuracy respectively.

Observe that this confirms our concerns: not only is k-WTA not an
effective defense, it makes adversarial training \emph{worse}.
This is not the first time that this has happened:
\citet{athalye2018obfuscated} previously observed the same effect of a defense \emph{reducing} the efficacy of adversarial training.

\lessons
\begin{myenumerate}
	\item Score-based or decision-based attacks can work as well or even better than gradient-based attacks, in particular when the gradients are (intentionally) masked.
	\item Transfer-based attacks, while cheap to implement, require
          a highly similar source model.
          The best transfer attacks typically succeed less than half of the time, and
          thus transfer attacks typically cannot disprove robustness claims less than
          roughly $50\%$.
\end{myenumerate}

\exercise
\begin{myenumerate}
  \item Instead of performing a black-box attack, is there a smooth and differentiable approximation of $k$-WTA that allows direct gradient descent to succeed?
  \item Increasing the sparsity ratio $\gamma$ reduces the fluctuations in the network activations. 
Is it possible to transfer adversarial examples from a less-sparse model to the sparser model, 
analogous to a prior attack on Stochastic Activation Pruning~\citep{dhillon2018stochastic, athalye2018obfuscated}?
\end{myenumerate}

\section{The Odds are Odd}
\label{sec:odds}

This paper~\citep{roth2019odds} proposes a statistical test for detecting adversarial examples, based on the distribution of a classifier's logit values (i.e., the per-class scores in the last layer of the classifier, before the softmax computation).

\howitworks

The main assumption that motivates the paper's defense is that adversarial examples are less robust to noise than benign examples.
More specifically, for a given input $x$, the defense compares the ``clean'' logit vector $\logits(x)$ to a ``noisy'' logit vector $\logits(x + \delta)$, where $\delta \sim \mathcal{N}$ is sampled at test-time from some fixed distribution $\mathcal{N}$ (e.g., a multinomial Gaussian).
The assumption is that for clean examples, $\logits(x) \approx \logits(x + \delta)$, while for adversarial examples the two logit vectors will differ significantly.

In more detail, the defense looks at variations in the difference between the logit of the predicted class $y$ and the logits of other classes $i \neq y$. Let
\begin{equation*}
\Delta_{y,i}(x)  = \logits(x)_i - \logits(x)_y \;,
\end{equation*}
where $y$ is the predicted class, $i \ne y$ is some other class, and $\logits(x)_j$ is the logit for class $j$.
Note that $\Delta_{y,i}(x)$ is always negative.
For each class, the defense checks whether these logit differences are robust to noise by computing\footnote{The values $\bar{\Delta}_{y,i}(x)$ are further normalized using means and standard deviations computed on the training set. This step is unimportant to understand the defense and our attacks.}
\begin{equation*}
\bar{\Delta}_{y,i}(x)  = \mathbb{E}_{\delta \sim \mathcal{N}}\left[\Delta_{y,i}(x+\delta) - \Delta_{y, i}(x)\right] \;.
\end{equation*}
In practice, the expectation is approximated by sampling $k$ independent noise values $\delta \sim \mathcal{N}$.
An input is rejected as adversarial if for any class $i$ other than the predicted class $y=f(x)$, we have $\bar{\Delta}_{y,i}(x) > \tau_{y, i}$, 
where the $\tau_{y, i}$ are thresholds set to achieve some fixed false-positive rate. 

\whyinteresting
The approach taken by this defense is very simple, but relies on a common misconception that the peculiarities of adversarial examples created by standard attacks can be used to detect \emph{all} adversarial examples. This defense is very similar in spirit to the one analyzed in \cref{sec:weakness}.\footnote{We break the defense in \cref{sec:weakness} with
  a single-purpose attack that is not as generalizable as the one described in this section,
  but which demonstrates the wide space of adaptive attacks.}

\hypothesis
The paper's evaluation shows that existing attacks such as PGD~\citep{madry2017towards} or C\&W~\citep{carlini2017towards} indeed produce adversarial examples with large values of $\bar{\Delta}_{y,i}(x)$. Due to the stochastic nature of the defense, the defense is further evaluated on an attack that combines PGD with ``Expectation Over Transformation'' (EOT)~\citep{athalye2017synthesizing} to produce adversarial examples robust to noise from $\mathcal{N}$.\footnote{A description of each of these attack techniques is in \cref{sec:attack_background}.} While more effective, this attack's success rate remains low.

The paper's evaluation shows no clear ``red-flags'', in part because very few results beyond the absolute detection numbers are reported (which perhaps could be considered a red-flag: it is helpful for defenses to give additional statistics if only to diagnose potential errors).

To better understand why existing attacks fail, and what a successful attack would have to achieve, we take a closer look at the detection statistics computed by the defense:
\begin{align*}
\bar{\Delta}_{y,i}(x) 
&= \mathbb{E}_{\delta \sim \mathcal{N}}\left[\Delta_{y,i}(x+\delta) - \Delta_{y, i}(x)\right] \\
&= \mathbb{E}_{\delta \sim \mathcal{N}}\left[(\logits(x+\delta)_i - \logits(x+\delta)_y) - (\logits(x)_i - \logits(x)_y)\right] \\
&=  \underbrace{\mathbb{E}_{\delta \sim \mathcal{N}}\left[(\logits(x+\delta)_i - \logits(x+\delta)_y)\right]}_{s_1}  + \underbrace{\left(\logits(x)_y - \logits(x)_i\right)}_{s_2} \;.
\end{align*}
We focus on the scores $s_1$ and $s_2$ independently. Observe that $s_1 + s_2$ must be small to evade the defense.

The term $s_1$ is positive and large if the model often switches its prediction to some other class when the input is perturbed with noise. The term $s_2$ is positive and large if the model is overly confident in its original prediction of class $y$, compared to other classes $i \ne y$. 

From here, we notice an issue with the type of attacks evaluated in the paper.
PGD creates examples with abnormally large values of $s_2$ (compared to benign examples) as this is precisely the objective that the attack maximizes. In contrast, attacks such as C\&W that find the smallest adversarial perturbation will find examples that lie very close to the decision boundary. For these, $s_2$ will be small, but $s_1$ will be large as examples close to the boundary are typically not robust to noise. Finally, combining PGD with EOT will reduce the term $s_1$, but will not prevent the attack from finding examples with abnormally large values of $s_2$.

Thus, a successful attack will be one that finds adversarial examples for which: (1) the model's robustness to noise is no smaller than for an average benign example; and (2) the model's confidence is no larger than for an average benign example. Interestingly, for a logistic regression model these two properties of resistance to noise and model confidence are directly related: the model's confidence depends on an input's distance to the decision boundary, which in turn defines the input's robustness to (Gaussian) noise.

\attack

We perform our evaluation on CIFAR-10, using the code released by the authors. We consider $\ell_\infty$ attacks bounded by $\epsilon = 8/255$, for which the paper claims a $71.4\%$ detection rate with a standard wide ResNet architecture. We consider two default settings proposed by the authors, one where the noise $\delta$ is sampled from a single multinomial Gaussian, and one where the terms $\bar{\Delta}_{y,i}(x)$ are computed for $10$ different sources of noise (Gaussian and uniform distributions with different variances). In the latter case, inference is extremely expensive: classifying a single example $x$ requires computing $z(x+\delta)$ for thousands of noise values $\delta$. 

As a first attack strategy, we create adversarial examples for which the term $s_2$ is the same order of magnitude as for benign samples. To do this we generate an adversarial example that targets the logits of an arbitrary benign example from another class. This is a ``feature-level'' attack~\citep{sabour2015adversarial}, where we aim to match the model's internal representation of another class, rather than just changing the predicted output. For each class $t$, we pick a benign example $x_t$ of that class and compute the logits $\logits(x_t)$. Then, for an example $x$ of class $y$, we pick a target $t \neq y$ and use PGD to create adversarial examples $x'$ that minimize the objective:
\begin{equation}
\label{eq:logit-match}
||\logits(x') - \logits(x_t)||_2^2 \;.
\end{equation}
This attack is surprisingly effective. The attack always fools the classifier, and only a small number of adversarial examples are detected due to abnormally large values $\bar{\Delta}_{y, i}(x)$. For the defense that uses a single noise source, $17\%$ of attacks are detected. For the defense that combines 10 noise sources (and therefore needs to lower the thresholds $\tau_{y, i}$ to achieve a fixed false-positive rate), only $10\%$ of attacks are detected.

To increase the attack strength, we further add the $s_1$ term to our objective. To minimize $s_1$, we apply standard EOT. Crucially, we apply EOT to the logit-matching objective in equation~\eqref{eq:logit-match}, rather than to the standard PGD objective. The resulting attack always fools the classifier maintaining a $0\%$ detection rate; lower than the false positive rate.

A very similar attack on this defense was independently presented by \citet{hosseini2019odds}.

\lessons

\begin{myenumerate}
	\item  Most existing attacks produce adversarial examples that are either abnormally close to the decision boundary (e.g., C\&W), or have abnormally high confidence (e.g., PGD). While this makes these attacks easy to detect, an adaptive attack can create examples that are similarly close to boundaries and have similar confidence as benign examples.
	\item This defense serves as a canonical example of the power of a very simple ``logit-matching'' attack that is effective against multiple defenses.
	\item When attacking a randomized defense, expectation-over-transformation (EOT) is an insufficient strategy if the attack \emph{objective} is not also appropriately adapted.
\end{myenumerate}

\exercise

It has been shown that for adversarially trained models, it is hard to perturb examples so as to match the logit representation of an example from a different class~\citep{engstrom2019adversarial}. Thus, the above logit-matching attack will probably not be effective if the defense is instantiated with a robust classifier. This leads to the following questions:
\begin{myenumerate}
	\item How effective is this paper's detection mechanism when instantiated with a robust classifier?
	\item Is there a better adaptive attack for this combined defense, for example by more directly incorporating the term $s_2$ into the attack objective?
\end{myenumerate}

\section{Are Generative Classifiers More Robust?}
\label{sec:generative}

This defense~\citep{li2018generative} is fairly complex and displays many characteristics of what makes an evaluation challenging, such as the use of multiple models, aggregation of multiple losses, stochasticity, and an extra detection step.

\howitworks

The defense is based on the Variational Auto-Encoder framework~\citep{kingma2013auto}. 
Specifically, the authors assume that data points $(x, y)$ are generated by some unknown process involving an unobserved latent variable $\eta$.\footnote{Our notation differs slightly from that in the paper, which denotes the latent variable by $z$, because we use $z$ for the logit vector.} The paper explores different ways of factoring the joint distribution $p(x, y, \eta)$. We focus on the ``GBZ'' model that the authors found to be most robust  (see~\cite{li2018generative} for a graphical illustration):
\[
p(x, y, \eta) = p(\eta)p(y|\eta)p(x|\eta) \;.
\]
Rather than delving into the theory of Variational Auto-Encoders, we present here the high-level classification algorithm implemented in the source code released by the authors (see \cref{alg:generative}). This mirrors our own mental model when assessing the robustness of this defense: understanding and evaluating what the implementation does is often more useful than understanding the theoretical or intuitive explanations made in the paper. 

The classifier combines three models, a per-class encoder $\texttt{enc}$ that produces parameters for sampling random latent vectors $\eta$, and two decoders $\texttt{dec}_1, \texttt{dec}_2$ that reconstruct the input and label from the latent vectors. All these models use standard CNN architectures. For each class, the defense runs these three models for $N$ random latent vectors ($N=10$ by default), and computes four scores that combine to form the logit for that class.

The paper proposes two extensions. First, to scale to CIFAR-10, the input $x$ is fed into a feature-extractor $\phi = \Phi(x)$, where $\Phi$ is an intermediate layer of a pre-trained VGG classifier. 
Second, the paper adds a detection step, that rejects inputs with ``unusual'' class probabilities $f(x)$ (obtained by applying the softmax function to the logits $\logits(x)$). We focus on the best-performing approach, where the detector rejects inputs with high KL-divergence between $f(x)$ and the mean probability vectors from training inputs of the same class.

\begin{algorithm}[t]
	\setstretch{1.1}
	\SetAlgoLined
	\KwIn{Data point $x$}
	\KwOut{Logits $\logits(x)$}
	\BlankLine
	\For(\tcp*[f]{iterate over class labels}){$k \in [1, K]$}{
		$\vec{\mu}, \vec{\sigma} = \texttt{enc}(x, k)$ \tcp*[f]{class-conditional encoder} \\
		\For{$i \in [1, N]$}{
			$\eta \gets \mathcal{N}(\vec{\mu}, \vec{\sigma} \cdot I)$ \tcp*[f]{sample latent vector}\\
			$x^* = \texttt{dec}_1(\eta)$ \tcp*[f]{decode to input}\\
			$y^* = \texttt{dec}_2(\eta)$ \tcp*[f]{decode to logits}\\
			$L_{\text{recons}} = ||x - x^*||_2^2$ \tcp*[f]{input loss}\\
			$L_{\text{CE}} = \text{cross-entropy}(y^*, k)$ \tcp*[f]{label loss}\\
			$\texttt{p}_{\text{prior}} = \log(\mathcal{N}(\eta; 0, I))$ \tcp*[f]{log prior}\\
			$\texttt{p}_{\text{posterior}} = \log(\mathcal{N}(\eta; \vec{\mu}, \vec{\sigma} \cdot I))$ \tcp*[f]{log posterior}\\
			$\texttt{score}_i = -L_{\text{recons}}  - L_{\text{CE}} + \texttt{p}_{\text{prior}} - \texttt{p}_{\text{posterior}}$
		}
		$l_k = \log \left( \frac1N \sum_{i=1}^N \exp({\texttt{score}_i}) \right)$ \tcp*[f]{logit}
	}
	\Return $\logits(x) = [l_1, \dots, l_K]$
	\caption{Generative Classifier~\citep{li2018generative}.}
	\label{alg:generative}
\end{algorithm}

\whyinteresting
The defense relies on a generative approach which has often been postulated as being more robust than purely discriminative models. This defense's complexity makes it particularly hard to evaluate and illustrates the advantage of focusing on simple attacks that precisely target the defense's most important pieces.

\hypothesis

Our initial observation is simply that this defense is \emph{complex}\footnote{Deep generative models tend to be much more complex than standard discriminative models, so the complexity of this defense is not unwarranted. But this does make a strong adaptive evaluation challenging.} (see \cref{sec:ecc} for another complex defense). It involves three networks, four scoring terms, many mixes of exponentials and logarithms, heavy randomization, and a detection step. It thus comes at little surprise that existing attacks (e.g., PGD~\citep{madry2017towards} or C\&W~\citep{carlini2017towards}) are shown to be ineffective in the paper: combining multiple loss terms effectively is never easy. The evaluation explicitly demonstrates the insufficiency of these attacks: 
attacks with unreasonably large $\ell_\infty$-bounds  ($\epsilon=0.5$ on MNIST, and $\epsilon=25/255$ on CIFAR-10) fail to bring the classifier's accuracy to $0\%$.

We set out to test the following initial hypotheses, informed by our reading of the paper and of the released code:

\begin{myenumerate}
	\item It is unlikely that the four terms that make up $\texttt{score}_i$ are all of similar importance. Targeting a single one might be sufficient and easier to optimize. 
	\item Given a successful attack on the classifier, it is plausible that only slight tweaks to the distribution of the logits $\logits(x)$ will be needed to also fool the detector.
\end{myenumerate}

We evaluate the paper's MNIST and CIFAR-10 models, using code and models provided by the authors. 
For simplicity, we exclusively consider a standard $\ell_\infty$ threat-model with $\epsilon=0.3$ on MNIST and $\epsilon=8/255$ on CIFAR-10 (the paper also gives results for other bounds and for $\ell_2$ attacks).

After a few standard yet ineffective tests (increasing PGD iterations and the number of random samples $N$), we tried a simple attack using random perturbations and many restarts.
This attack was surprisingly effective: after a few thousand trials the MNIST model's accuracy dropped below $50\%$. While trying to build a stronger attack (see details below), we noticed a curious fact about this defense: the success of our random attack is not due to the model's lack of noise-robustness, but to the high variance in the model's estimate of the $\texttt{score}_i$ values. Indeed, even when classifying an \emph{unperturbed} input drawn from the test set multiple times in a row, the model misclassifies it with some small ($5\%$-$10\%$) probability. The defense is thus vulnerable to a ``trivial'' attack that continuously submits the same input until the model misclassifies it by chance. This can be fixed by increasing the number $N$ of sampled latent vectors, at a cost of more expensive inference.

We thus searched for an attack that could produce more ``robust'' adversarial examples. 
To test our first hypothesis, we compared the different terms in the scores $\texttt{score}_y$ and $\texttt{score}_{i}$, where $y$ is the true class and $i \neq y$ some other class. We found that $L_{\text{CE}}$ is the only important term. The other terms vary little across class scores. 
To generate more robust attack samples, we also increase the defense's parameter $K$, the number of randomly sampled latent vectors from $10$ to $100$ (this is essentially the EOT technique~\cite{athalye2017synthesizing, athalye2018obfuscated}). The created samples are tested on a defense with the original parameter setting of $K=10$. With a standard PGD attack to maximize the $L_{\text{CE}}$ term for the true class, we reduce the classifier's accuracy below $5\%$ on both MNIST and CIFAR-10. These adversarial examples are also robust to the model's high variance.

We found that despite only targeting the classifier, this attack is moderately successful against the KL-divergence detector: only $20\%$ of the MNIST attacks and $10\%$ of the CIFAR-10 attacks are detected. This gives credence to our second hypothesis. Finally, we proceed to find adaptive strategies that fool both the classifier and detector. 

\attack
We first directly minimized the KL-divergence, which proved computationally expensive (as it requires optimizing over all class scores) and unsuccessful. We further tried to align the model's decoded logits $y^* = \texttt{dec}_2(\eta)$ to those produced on a clean input from a different class, using the ``feature-adversary''~\citep{sabour2015adversarial}. The aim is to match the model's internal representations to those obtained with another input, so that downstream statistics (e.g., the softmax output) match as well.
On CIFAR-10, this attack achieves $99\%$ success against the combined classifier and detector.

The insight of feature adversaries suggests a much simpler attack we had overlooked. As the CIFAR-10 classifier is of the form $f(\Phi(x))$ where $\Phi(x)$ are features extracted from a VGG model, an attack that transforms these features to those of a different class will be successful.
Concretely, we find an adversarial example $x'$ that minimizes $L(x') = ||\Phi(x') - \Phi(x_{t})||_2^2$, where $\Phi(x_{t})$ are the VGG features for a clean example $x_t$ of a different class. This attack succeeds with $100\%$ probability.

For the MNIST classifier, the same attacks as above produce adversarial examples that each succeed with $90\%$ probability (over the model's internal randomness). This limitation may be inherent as the model also has a $10\%$ failure rate when classifying a clean input. To verify that our attack's ``failure'' is indeed due to the classifier's excessive variance, we made the sampling of latent vectors $\eta$ deterministic (i.e., given parameters $\vec{\mu}, \vec{\sigma}$, the samples $\eta_1, \dots, \eta_N$ are produced using a fixed seed). In this simplified setting (which has no effect on the model's accuracy), we reach $100\%$ success rate with the following PGD attack: 
we maximize $L_{\text{CE}}$ for the true class, and stop as soon as we find 
that the perturbation computed by the current PGD iteration also fools the detector.

\lessons

\begin{myenumerate}
	\item For complex defenses, it is useful to decompose the contributions of individual classification steps, in order to determine which ones should be focused on.
	\item Evaluating a defense on random noise can reveal surprising failure cases.
	\item Feature adversaries are useful for jointly evading a classifier and detector, as they ensure that any statistics computed on top of adversarial features match those from clean examples.
\end{myenumerate}

\exercise

\begin{myenumerate}
	\item Do there exist more robust adversarial examples on the MNIST classifier, that succeed with probability close to $100\%$ over the classifier's randomness?
	\item Do some of the attacks described above translate to other generative classifiers ~\citep{schott2018towards,willetts2019disentangling}?\end{myenumerate}

\section{Robust Sparse Fourier Transform}
\label{sec:fourier}

This paper \citep{bafna2018thwarting} introduces a defense to $\ell_0$
adversarial examples through what is called a ``robust sparse
Fourier transform''.

\howitworks

For a given input, this paper proposes to defend against $\ell_0$
adversarial examples by taking each image, ``compressing'' it by
projecting to the top-$k$ coefficients of the discrete cosine
transform, inverting that to recover an approximate image,
and then classifying the recovered image.
The classifier is trained on images compressed in this way so
that it remains accurate.

In more detail, this paper uses the Iterative Hard Thresholding
(IHT) method which consists of $T$ iterations, each of which
performs a pass at compressing the input $x$ with a Fourier
transform and then recovering the inverted input. 
This defense's idea is  conceptually similar to that of the ME-Net 
defense studied in Section~\ref{sec:menet}, although ME-Net 
has the additional challenge of being randomized.

\whyinteresting
There are not many defenses to $\ell_0$ adversarial examples, and
applying a Fourier transform intuitively makes sense as a
possible defense.

\hypothesis

This paper does not contain an analysis of its robustness 
to adaptive attacks.
Instead, it demonstrates that adversarial examples on a base
classifier, when pre-processed by the defense method, are
no longer classified as adversarial by the base classifier
afterwards.
This is insufficient to demonstrate robustness as has been
argued extensively \citep{carlini2019evaluating}.

In particular, the paper claims over $75\%$ accuracy
at an $\ell_0$ robustness of $55$ on MNIST; however,
it has been shown that an $\ell_0$ distortion of $25$ is
sufficient to actually cause humans to change their
classification of MNIST images~\citep{jacobsen2019exploiting}: it is therefore
improbable that any defense could achieve a stronger result.

We suspected that an adaptive attack would break this
defense without any other difficulty.
The paper argues robustness on multiple $\ell_0$ attacks,
including the one presented by \citet{carlini2017towards}
which often has lower distortion than the others considered.
We therefore adapt this attack to the defense.

\attack

To begin we implement this defense as a differentiable
pre-processor in front of the neural network classifier
(using code provided by the authors).
Implementing the procedure as a differentiable function
requires a small amount of work as there are no
non-differentiable components: everything is already
fully differentiable.
We then apply the $\ell_0$ attack from \citet{carlini2017towards}
on this combined function without any further modification.
We run the attack with a maximum of $500$ iterations of gradient 
descent before deciding which pixels to freeze at their original
value.
Instead of running a multi-targeted attack we run the attack in
an untargeted manner, resulting in a larger distortion than we
could otherwise achieve but $10\times$ faster and
mirroring the original evaluation.
We find the resulting attack is successful at generating
adversarial examples with a median distortion of
$14.8$ pixels, within the margin of error of the $15$ pixels
reported by \citet{carlini2017towards} for an undefended model.

\lessons
\begin{myenumerate}
   \item It is important to optimize the combined loss function on  $f(p(x))$ whenever defenses
	propose a pre-processing function $p$.
	\item Not all pre-processing functions are hard to differentiate. Some just require to be implemented in a library that supports auto-differentiation.
\end{myenumerate}

\exercise
\begin{enumerate}
\item The pre-processing method is differentiable: at each
  iteration of the IHT procedure, the inner computation is
  a floating point value.
  Can the defense be broken if we artificially make it non-differentiable, 
  e.g., by quantizing the output of IHT?
  
\item To make models robust to $\ell_0$-attacks, one option is to train a model against $\ell_1$-attacks~\citep{tramer2019adversarial}. Is there an approach that is specifically aimed at the $\ell_0$ norm?
\end{enumerate}

\section{Rethinking Softmax Cross Entropy}
\label{sec:rethinking}

This paper \cite{pang2020rethinking} proposes a new loss function to use during training
in order to increase adversarial robustness.

\howitworks

The paper trains a feature-extractor $g$. Note that for this classifier, the output features $g(x) \in \mathbb{R}^N$ do not correspond to class-scores.
Instead of training models with standard softmax cross entropy, this paper
proposes the \emph{Max-Mahalanobis center (MMC) loss}, defined as
\[
L_{\text{MMC}}(g(x), y) = \frac12 \| g(x) - \mu_y\|_2 \;.
\]
Here, the $\mu_y \in \mathbb{R}^N$ are the centroids of the Max-Mahalanobis distribution;
the exact choice of $\mu$ is not important for understanding the defense. Given the trained feature-extraction network function $g$, the network classifies an input $x$ into one of $K$ classes as
\[f(x) = \text{arg min}_{1 \leq i \leq K} \lVert g(x) - \mu_i \rVert_2.\]

We comment on one further observation we make of this paper.
While not a failure mode, this paper contains a large amount of theory and proofs of
various facts that are mostly unrelated to the actual defense proposal itself.
While this theory may be interesting in its own right, it is not always necessary to
follow the theory for \emph{why} the defense was proposed in order to still evaluate
 the proposed defense.

\whyinteresting

There are a number of defenses that aim to replace the standard softmax cross entropy
loss with alternate functions.
We select this defense as as a representative example of this style of defense because it is a
strengthening of a prior defense to adversarial examples \citep{pang2018max}, and as such
we expected it would be a strong candidate.

\hypothesis

This defense does not perform an analysis of robustness to adaptive attacks; 
as such, we expect that carefully constructing a loss function that is well
suited to the defense will allow us to reduce the accuracy significantly.

In general,
whenever a defense changes the loss function used to train a model, our first hypothesis
is always that the new model has a loss surface that is not well suited to attacks
with the standard softmax cross entropy loss.
Towards that end, we perform the analysis that we would have liked to see in the original paper.

The original evaluation uses standard attacks such as PGD that maximize the cross-entropy loss over the model's predictions. The paper defines the defense's logit vector $z(x)$ as $z(x)_i = -\|g(x)-\mu_i\|_2$ for $1 \leq i \leq K$, and then applies the softmax function to $z(x)$ to obtain class probabilities.
Initially we examined the value of the logits of the classifier and found that they
were extremely abnormal compared to a typical classifier.
The largest logit tended to be near $0$ and the remaining logits near $-200$.
This reminded us of distillation as a defense \cite{papernot2016distillation}, which has
a similar effect on the logits.
A simple method for breaking distillation is to simply divide the logits by a large
constant \cite{carlini2016defensive}; however when we tried this it was not sufficient
to break the defense completely (although it did help some and reduced the classifier
accuracy on CIFAR-10 from $24\%$ to $18\%$.
That this alone allowed the attacks to succeed more often was worrying: shifting the
logits by a constant should not improve the efficacy of attacks.

As such, we hypothesized that an attack would need to design a loss function that
better captured what the defense was doing in order to succeed more often.

\attack

The justification for what the defense does, as well as its training procedure,
are complicated. Yet, inference is simple.
Given the trained feature-extraction network function $g$, the network assigns the label
\[\text{arg min}_i \lVert g(x) - \mu_i \rVert_2.\]

That is, there are $K$ (for a $K$-class classification problem)
different ``target'' vectors $\{\mu_i\}_{i=1}^K$ and the classifier
returns whichever target is closest given the feature vector $g(x)$.
In this sense, this defense is similar to the error correcting defense from
Section~\ref{sec:ecc}.

This gives us our candidate attack procedure.
Instead of indirectly performing softmax cross entropy loss on the distances to $\mu$,
we directly target each of the target vectors $\mu$.
That is, we directly optimize the loss function
\[L(x) = \lVert g(x) - \mu_i \rVert_2^2\]
for each target class $i$, again performing a multi-targeted attack.
Directly optimizing the loss function in this way reduces the classifier accuracy to
under $0.5\%$ at distortion $\epsilon=0.031$ on CIFAR-10.

\lessons

\begin{myenumerate}

\item Any time a defense changes the loss function used during training, the
loss function used to attack should always be studied carefully.

\item The softmax cross-entropy loss is hard to optimize when there are large differences in the model's logit values.
  
\end{myenumerate}

\exercise
\begin{myenumerate}
	\item The logit margin loss introduced by~\citet{carlini2017towards} can outperform the cross-entropy loss for abnormal logit values. Can applying PGD with this loss reduce the defense's accuracy towards $0\%$?
\end{myenumerate}

\section{Error Correcting Codes}
\label{sec:ecc}

This paper \citep{verma2019error} proposes a method for training an ensemble of models with
sufficient diversity that the redundancy can act as error correcting codes
to allow for robust classification.

\howitworks

Given an $K$-class classification problem, this paper proposes to generate a matrix $A \in \{0, 1\}^{K \times M}$ of binary codewords.\footnote{The paper generates the matrix $A$ using some procedure that is unimportant for
understanding the defense.}
The defense trains $M$ different classifiers
so that classifier $f_i$ outputs a value between $0$ and $1$ with the objective
that $f_i(x) = A_{j,i}$ when the label for $x$ is $j$.
This way, each classifier is performing a binary task prediction where each class
is randomly assigned to either the 0 or the 1 class.
Clearly if there are not enough classifiers, we can not uniquely recover the
result given the output of the models (in particular, we must at least have that $M \ge \log(K)$).
Thus, each classifier $f_i$ is trained on a slightly different task, aiming to increase
diversity of the classifiers so that adversarial examples will not transfer between
the classifiers.

Specifically, each classifier is trained as a function $z_i \colon \mathcal{X} \to \mathbb{R}$
and then $f_i(x) = \text{sigmoid}(z_i(x))$.
To generate the final predictions, define the vector
 $Z(x) = \begin{bmatrix}z_1(x) & z_1(x) & \dots & z_M(x)\end{bmatrix}$,
and then return $f(x) = \text{arg max}_{1 \leq j \leq K} \lVert \text{sigmoid}(Z(x)) - A_j \rVert_2$.

\whyinteresting

We study this defense, along with the the Ensemble Diversity \citep{pang2019improving} defense in the prior section,
as representative examples of defenses that ensemble together multiple models.
Because the prior defense was not robust, we selected another ensembling defense to
test if it would be any more robust to adversarial examples.

\hypothesis

This is another defense that we would classify as \emph{complex}, even though
it may not look that way initially.
It is not \emph{difficult}---it is an easy defense to understand, conceptually---but
there are several moving pieces where multiple classifiers are trained independently and
each classifier is trained on a different task.
As we saw previously with the generative model defense~\citep{li2018generative} in \cref{sec:generative},
attacks on complex defenses do not have to be complex once the weak links are identified.

Our main concern with this paper was that it reports nearly $40\%$ accuracy at an $\ell_\infty$
distortion of $\varepsilon=0.5$ which is a tell-tale sign of gradient masking \citep{athalye2018obfuscated}:
any defense at this distortion \emph{must} be breaking the gradient descent process somehow.
This paper draws exactly the opposite conclusion from this figure, stating
that ``Because model accuracy rapidly drops to near 0 as $\varepsilon$ grows,
the [accuracy versus distortion] figure provides crucial evidence
that our approach has genuine robustness to adversarial attack and is not relying on ``gradient-masking''.''
While it is true that the accuracy does go down, it does not reach $0\%$ accuracy at
$\varepsilon=0.5$ and therefore the defense \textbf{is} causing gradient masking.

When we study the code we find the reason for this.
Each of the classifiers $f_i$ are followed by a sigmoid activation function in
order to map to the correct output space for the 0-1 codes.
However, after this, to compute the unified predictions, the implementation takes
the $\log$ of the result and then feeds this into the $\text{softmax}$ function.
Finally, to generate adversarial examples with the cross-entropy loss, another
$\log$ is required.
This is exceptionally numerically unstable and we suspected it was the cause
of the gradient problems.

\attack

Given the $M$ neural networks $\{f_i\}_{i=1}^M$, we remove the final sigmoid layer
after each model, and remove the $\log$ operation in merging together the predictions.
In order to be more robust to further numerical instabilities, we replace the standard
softmax cross entropy with the hinge loss proposed by \citep{carlini2017towards}.
This attack is able to substantially degrade the model accuracy to lower than $20\%$
on CIFAR-10 at $\varepsilon=0.031$, however it does not completely reduce it to zero.

We then apply a few tricks that are typically able to reduce the accuracy by a few
percentage points more.
First, we run a multi-targeted attack and generate untargeted adversarial examples by
targeting each of the other 9 target labels.
Second, we run each of these multi-targeted attacks five times, with different initial random 
steps for the PGD attack.
Finally, if  \emph{any} inner iteration of PGD succeeds at generating an adversarial
example we take this; we don't require that the \emph{final} output of PGD after
exactly 100 steps is an adversarial example.
This finally brings the accuracy of the classifier to under $5\%$ at $\varepsilon=0.031$,
compared to the reported $57\%$ accuracy at this distortion.

\lessons

\begin{enumerate}
\item Combining together multiple neural networks into a single combined
  defended network is extremely error-prone; however, once a simple method
  is identified, it often is effective.
\item There are number of tricks that can increase attack success rate by
  a few percentage points each; when model accuracy is already low, these tricks
  can be sufficient to reduce accuracy to near-zero.
\end{enumerate}

\exercise

\begin{enumerate}
  \item Suppose that instead of returning the closest code word given the output of the
model, the defense only returned the class if \emph{all} codeword bits agreed (and
rejected the input otherwise). Clearly
this will reduce classifier accuracy---but will it increase robustness?
\item Is there a different loss function for PGD, so that an untargeted attack with a single restart reduces the model's accuracy below $5\%$.
\end{enumerate}

\section{Ensemble Diversity}
\label{sec:ensemble}

This paper~\citep{pang2019improving} proposes to train an ensemble of models with an additional regularization term that encourages diversity. Compared to single models or less diverse ensembles, this additional diversity is supposed to make it more difficult for attacks to find minimal adversarial examples.

\howitworks

Let $f_m(x)$ be the probability vector of the $m$-th model in the ensemble and $f(x) = \sum_m f_m(x)$ be the probability vector output by the ensemble. The ensemble of models is trained on the following objective,
\begin{align*}
	{L}(x, y) = - \alpha \mathcal{H}(f(x)) - \beta \operatorname{Vol}^2\left(\left\{f_m^{\backslash y}(x)\right\}\right) + \sum_{m}^{M} L_{CE}(f_m(x), y)
\end{align*}
where $\mathcal{H}(\cdot)$ is the Shannon entropy, $L_{CE}$ is the standard cross-entropy loss applied to the prediction of each model and the middle term is the volume spanned by the probability vectors of the individual models of the ensemble (each normalized to unit length). Note that the volume is only computed over the non-maximal predictions, i.e. we remove the leading class from each probability vector $f_m(x)$. 
The rational for doing this is that this will
balance accuracy and diversity (the volume can only increase 
if the predictions in the non-maximal classes are diverse).
The weighting coefficients $\alpha, \beta$ are hyper-parameters. 

\whyinteresting

There is a long line of work studying the robustness of neural network ensembles.
Most other work in this direction has not succeeded \citep{he2017adversarial}.
We study this defense, along with that of \citet{verma2019error} (\cref{sec:ecc}) and \citet{pang2019improving} (\cref{sec:empir}), to evaluate what appears to be the strongest defenses in this class.

\hypothesis

None of the terms in the loss function seem to encourage any kind of gradient masking. There is no other mechanism other than the change of the training objective, thus suggesting that standard off-the-shelve gradient-based attacks should be successful.

The paper's results section might impress with the large number of attacks employed against the ensemble model. 
However, BIM~\citep{kurakin2016adversarial}, PGD~\citep{madry2017towards} and MIM~\citep{dong2018boosting} are extremely similar and if one of these attacks fails the others are likely to fail as well. Instead, it is better to employ a more diverse set of attacks, e.g. by including the score-based pixel-wise attack \citep{lukas_abs} or the Brendel \& Bethge (B\&B) attack~\citep{brendel2019bethge} (the latter of which was not available at the time of initial publication). Similarly, the results for FGSM are only useful in that they might signal gradient masking (in particular if FGSM finds adversarial perturbations where multi-step methods do not) but in general they should always perform worse then more powerful attacks (which they do here).

Nonetheless, the paper's results contains a few odd values. For one, the diverse ensemble reaches higher robustness on CIFAR-10 than a state-of-the-art adversarially trained model (30.4\% compared to 27.8\% for $\epsilon = 0.02$). That is fairly unlikely and not mentioned anywhere in the main text. Second, it is odd that the BIM attack tends to be more effective than PGD given that the two only differ in their starting points (BIM starts from the original sample while PGD starts from a random point within the allowed $\ell_\infty$ norm ball). For adversarial training, choosing a random starting point can mitigate some gradient masking issues. In general, choosing a random starting point makes the attack converge to different perturbations which can increase success rate by 10\% or more if each sample is attacked multiple times. 

These oddities suggest that the attacks might not have converged. And indeed, the paper reports to have used only 10 iterations for BIM, PGD and MIM (with a step size of $\epsilon / 10$). Such a small number of iterations is unlikely to allow the attacks to converge and to find the strongest adversarial perturbation, suggesting that the attack success can be substantially increased simply by increasing the number of iterations. The same observation holds for the C\&W~\citep{carlini2017towards} and EAD~\citep{chen2018ead} attacks for which the paper reports $1{,}000$ iterations. Due to the inner hyper-parameter optimization that both attacks employ, this number is still on the low side.

Taken together, our hypothesis was that simply increasing the number of iterations and/or combining PGD with a substantially different attack such as B\&B can substantially increase attack success.

\attack

For our experiments, we evaluate the authors' released defense on CIFAR-10
with $\ell_\infty$-bounded attacks of distortion $\epsilon = 0.01$. We evaluate the strongest ensemble trained with $\alpha=2$ and $\beta=0.5$ for which the paper reports an accuracy of 48.4\%. Each model of the ensemble is based on the ResNet-20 architecture.

We started from the evaluation code as well as the pre-trained model weights provided by the authors. Simply increasing the step size by a factor of three reduces the accuracy from 48\% to 26\%. Increasing the number of iterations to 50 decreases the accuracy to 20\% while 250 iterations reduced the accuracy to 10\%. A further increase did not change the result, suggesting that 250 iterations are sufficient for convergence.

We then repeated the attack three times for each example, which decreased accuracy to 7\% for 250 iterations. The remaining samples seemed to be difficult for PGD, which is why we chose the $\ell_\infty$ version of the B\&B attack~\citep{brendel2019bethge}. In contrast to PGD, which starts at a point close to the clean sample and moves outwards to the worst-case adversarial within the $\epsilon$-ball, the B\&B attack starts from a misclassified sample far away from the clean sample and then moves along the boundary of adversarial and non-adversarial images towards the clean sample. As initial points for the B\&B attack, we chose large-perturbation adversarial examples ($\epsilon = 0.15$) generated by PGD with 20 steps. We then applied B\&B with 20 steps. This was sufficient to reach 5\% accuracy. Repeating the attack a few times on the remaining samples reduced the accuracy to 0\%.

\lessons

\begin{myenumerate}
	\item It is important to make sure that attacks actually converged and that the right
	hyper-parameters have been chosen. Also, if an attack uses random starting points, 
	the attack should be repeated several times on each sample.
	\item Using very similar attacks such as BIM, PGD or MIM is unlikely to yield very different
	results. Instead, attacks that use substantially different
	strategies such as the B\&B attacks~\citep{brendel2019bethge} should be considered.
\end{myenumerate}

\exercise

\begin{myenumerate}
	\item The optimal step size and the necessary number of iterations substantially differ between attacks but also between different defenses. Compare a few attacks (e.g. PGD, C\&W, B\&B as well as some score- and decision-based attacks). How many steps does each attack need until convergence? How sensitive are the results to the attack's hyper-parameters (e.g. the step size)?
	\item Repeating attacks a few times can substantially lower model accuracy. Which attacks can profit most from repetitions?
\end{myenumerate}

\section{EMPIR}
\label{sec:empir}

This paper \citep{sen2020empir} is the third defense we study that
develops a method to defend
against adversarial examples by constructing an ensemble of models,
but this time with mixed precision of weights and activations.

\howitworks

This paper trains multiple models $\{f_i\}$ with different levels of precision
on the weights and on the activations for each and returns the output as the
majority vote of the models.
Concretely, for a CIFAR-10 defense the paper uses one full-precision model, one model
trained with $2$-bit activations and $4$-bit weights,
and another model with $2$-bit activations and $2$-bit weights.

\whyinteresting
The prior two ensemble defenses we studied were not effective;
however this defense takes a different approach by training low-precision
models.

\hypothesis

This paper does not contain an explicit analysis of its robustness
to adaptive attacks, but performs an analysis of the defense
to standard gradient-based attacks on the ensemble.
However, given that the defense has potentially non-differentiable layers
we believed that it may be possible for various forms of gradient masking to occur.
Investigating the code, we found that the paper \emph{does} include a
BPDA-style estimate of the gradient of the backward pass.
Thus, we expected that performing gradient descent with respect to
the entire model ensemble and carefully ensuring that gradient
information was correctly propagated through the model
would suffice to bypass this defense.

\attack

We begin by constructing a loss function that we will use to generate
adversarial examples.
As a first attempt at an attack, we form the simplest loss function that we
could imagine: take the
class probability vectors of the three models $f_1(x), f_2(x), f_3(x)$ and
average them component-wise so that the prediction of our agglomerated model
for class $i$ is given by
$\hat{f}(x)_i = {\frac13}(f_1(x)_i + f_2(x)_i + f_3(x)_i)$.
Then we perform PGD on the cross-entropy loss for this model $\hat{f}$.
Notice that this loss function \emph{is not consistent}, violating one
of our lessons: because the classifier takes the majority vote in order to
decide the final prediction, only two of the classifiers must agree on the
target class; the third model could have $0$ confidence.
However, before we spend any extra effort developing a stronger loss
function we test this simple loss function.

Surprisingly, we find that it is effective. By running $100$ iterations
of PGD on this loss function, we are able to reduce the accuracy 
of the defense to $1.5\%$ at $\varepsilon=0.031$ (the original evaluation
reported accuracy of $13.5\%$ at $\varepsilon=0.1$ with $40$ iterations of PGD).

\lessons
\begin{myenumerate}
\item The attack themes we distill in \cref{sec:overview} are not hard-and-fast rules: it is possible to evade
  defenses while also (knowingly) disregarding them.
\item Ensembles of weak defenses still appear weak \cite{he2017adversarial}.
\end{myenumerate}

\exercise
\begin{enumerate}
\item We only construct a very simple loss function that reduces model accuracy
  to $1.5\%$. Can a stronger (probably consistent) loss function reduce the
  accuracy to $0\%$?
  At $\varepsilon=4/255$?
  
\item Try to attack each of the models $f_i$ individually. Are they all similarly robust, or is one harder to attack than the others? (If the robustness of the models differ,
  how might a more consistent loss function look to attack the full defense).
\end{enumerate}

\section{Temporal Dependency}
\label{sec:temporal}

This defense~\citep{yang2018characterizing} detects adversarial examples for automatic
speech recognition.

\howitworks

Given an audio sample $x \in [-1,1]^{t}$ represented as the raw audio waveform,
an automatic speech
recognition system generates a transcription of the input as $y \in \mathcal{Y}^n$
where $\mathcal{Y}$ is the output space of tokens---in
this paper, a character ``a'' through ``z'' or the whitespace
token.

In order to detect if a given input is adversarial,
the detection procedure checks if the classification of the
first half of the audio waveform is similar to the first half of the
classification of the complete audio waveform.
Formally,
let $s_{1..k}$ denote the prefix of length $k$ for
a sequence $s$, 
and let $\text{len}(s)$ return the length of a sequence.
The defense begins by setting $x_{\text{pre}} = x_{1..k}$
(the method for selecting $k$ will be discussed later).
Then, compute the predicted
classification $y = f(x)$ of the full audio waveform $x$
and $y_{\text{pre}} = f(x_{\text{pre}})$ as the prediction of the length-$k$ prefix of $x$.
From here, let $l = \text{len}(y_{\text{pre}} )$ and compute
$\phi = \text{sim}(y_{1..l}, y_{\text{pre}} )$
where $\text{sim}$ computes the similarity between $y_{1..l}$ and $y_{\text{pre}} $
according to some metric.
If $\phi$ is large then the audio is determined to be adversarial;
otherwise if $\phi$ is small then the audio is said to be benign.

In order to compute the similarity, the paper proposes various metrics
$\text{sim}$; because they all perform roughly the same, we choose the
simplest to evaluate: the character error rate.
That is, we compute the Levenshtein (edit) distance between the two
strings.
From source code we obtained from the authors we found that the only
detail in the implementation is that the edit distance omits whitespace
characters before computing the distance.

It only remains to discuss how to compute $k$.
The majority of
the paper considers the case where $k=t/2$, so $x_{\text{pre}}$ is the first
half of the audio sample.
The authors additionally study the cases were $k$ is selected either
from some small set of values, or where $k$ is sampled uniformly at
random from $0.2$ to $0.8$.
In order to attack the most difficult version of the defense we
fool the full randomized version of the detector.

\whyinteresting
This is the only defense that we have seen at ICLR, ICML or NeurIPS that is not focused on image classification; we therefore consider
it an interesting case study.

\hypothesis

The paper performs an extensive adaptive attack evaluation, trying
three different attack techniques.
One of these attacks looked almost exactly like what we believe would be
the correct attack:
select a target sequence $t$, and let $t_{\text{pre}}$ denote the prefix of $t$. Then, set the loss function to minimize the loss on both the transcription of the prefixed input
$f(x_{\text{pre}})$ and on the full input $f(x)$,
where the loss on the prefixed input is determined by the loss between the classification
of $f(x_{\text{pre}})$ and the
prefix of the target sequence $t$.
We noticed one potential failure mode.
First, a single hyper-parameter controls the relative importance of both
loss terms: the minimized loss function is:
\[\lVert \delta \rVert_2 + \lambda \cdot \big(L(f(x'),t) + L(f(x'_{\text{pre}}), t_{\text{pre}})\big) \;,\] 
instead of what we would have expected:
\[\lVert \delta \rVert_2 + \lambda_1 \cdot L(f(x'),t) + \lambda_2 \cdot L(f(x'_{\text{pre}}), t_{\text{pre}}).\]

We were confident that there was \emph{some} error in the evaluation because
the success rate of the attack is
low for an \emph{unbounded} attack:
even when never placing a bound on the total distortion the adversarial is
allowed to make, the attack does not succeed $100\%$ of the time at fooling
the detector.
Such an unbounded attacker should always \emph{eventually} succeed,
if only by adding so much noise that the original audio is completely
unrecognizable.

Our guess as to why this may happen comes from our experience
with the original attack codebase, which is \emph{not} a simple attack
and can thus be hard to debug.
Because the objective of the attack is to generate adversarial
examples of minimal distortion, the default attack is implemented
as follows.
First, choose an arbitrary distortion bound $\tau$ that is sufficiently
large.
Then, perform PGD~\cite{madry2017towards} on this distortion-bounded input space.
Once the attack is successful at fooling the classifier,
the distortion bound is reduced by 10\% and
the attack repeats.
There are a number of details here which could have complicated the
attack process, however because the authors do not release source code
for the attack we are unable to confirm our hypotheses.

\attack

We implemented the attack as described in the paper, beginning with just
one hyper-parameter $\lambda$---we only wanted to introduce a new hyper-parameter
if it turned out to be necessary.
Specifically, we perform gradient descent on the loss function
$L(x') = L(f(x'), t) + L(f(x'_{\text{pre}}), t_{\text{pre}})$
where $t_{\text{pre}}$ is computed as the first $l$ characters of the
\emph{true} prediction $t$ where $l=\text{len}(f(x'_{\text{pre}}))$.
Importantly, this is \emph{not} just the first $l$ characters
of $f(x')$: this value might not be the correct output at any point
in time.
If we try to send $f(x'_{\text{pre}})$ towards the current prediction $f(x')$
the two loss functions will be in conflict with each other.
However, because we know that $f(x')$ will eventually equal $t$,
we send $f(x'_{\text{pre}})$ to the prefix of $t$.
Once both the prediction of the model reaches our desired target
phrase and the detector is fooled completely (i.e.,  $f(x'_{\text{pre}}) = y_{\text{pre}}$
exactly) we reduce the distortion bound and continue our search.

In order to ensure that our adversarial examples remain effective
under random choices of the prefix length $k$ we run our attack by,
in parallel, ensuring that the attack is successful with $k=.25$,
$k=.5$ and $k=.75$ all simultaneously.
We hope from here that this implies robustness to values in between
(and empirically we find it does).

There are two important differences between this evaluation and all
other evaluations in this paper.
First, we perform a targeted attack here: untargeted attacks are
uninteresting on audio classification because they typically
just induce misspellings.
Second, because we run an unbounded attack, instead of reporting
the success rate at a given distortion bound, we report the
median distance to generate a successful perturbation.
Success rates for unbounded attacks should always be $100\%$ and
are not valid points of comparison:
given sufficient distortion, \emph{eventually}
we will succeed.
Indeed, we succeed in generating adversarial examples successfully that
reach the target class $100\%$ of the time, and that fool the detector
$100\%$ of the time.

On a baseline classifier, we require a median $\ell_\infty$ distortion
of $41$ (out of a 16-bit integer range of $65536$).
For the classifier defended with this defense, we require a distortion
of $46$, a slight increase but not significantly higher than
baseline defenses (e.g., bit quantization).

\lessons

\begin{myenumerate}
\item Subtle differences in attack implementations can distinguish
  between effective and ineffective attacks.
\item Defenses that do not rely on principled inclusion of randomness
  rarely benefit from its addition.
\end{myenumerate}

\exercise

\begin{myenumerate}
	\item
The proposed defense checks that $f(x_{\text{pre}}) = y_{\text{pre}}$.
Instead consider an alternate defense that splits the audio in two
two halves, the left half $x_l$ and the right half $x_r$;
the detection mechanism checks whether or not
$f(x) = f(x_l) \mid \mid f(x_r)$ where $\mid\mid$ denotes string
concatenation.
Is this defense effective?

\item What is the median $\ell_\infty$ distortion required to fool the model into outputting an empty transcription (i.e., a sequence of whitespace characters)? 
\end{myenumerate}

\section{Mixup Inference}
\label{sec:mixup}

This paper \cite{pang2020mixup} employs a stochastic interpolation procedure during inference in order to mitigate the effect of adversarial perturbations.

\howitworks

The defense works as follows. For each input $x$ (e.g. a clean or perturbed image) we compute $K$ interpolations with samples $s_{k}$,
\begin{equation*}
	\tilde x_k = \lambda x + (1 - \lambda) s_k.
\end{equation*}
where $\lambda$ is a fixed hyper-parameter ($\lambda = 0.6$ in all experiments) and $s_k$ is sampled randomly from a predefined set of images $\mathcal{S}$. We then average the logit responses $z(\cdot)$ of the undefended model over all $K$ interpolations, i.e. the final response $\hat z(\cdot)$ of the defended model is
\begin{equation*}
	\hat z(x) = \frac{1}{k}\sum_{k=1}^{K} z(\tilde x_k) = \frac{1}{k}\sum_{k=1}^{K} z(\lambda x + (1 - \lambda) s_k).
\end{equation*}
The paper develops two instantiations of the defenses: OL (other label) and PL (predicted label). The two types only differ in how $s_k$ is sampled from $\mathcal{S}$. For OL we sample $s_k$ uniformly from all images for which the predicted label is \emph{different} from $x$. For PL we sample $s_k$ uniformly from all images for which the predicted label is the \emph{same} as for $x$. The defense is applied with $K=1$ during training and with $K=15$ (OL) or $K=5$ (PL) during testing.

The motivation for this defense is two-fold: for one, the effect of a perturbation $\delta$ is reduced by $\lambda$ due to the interpolation. Second, the manuscript hypothesizes that a perturbation $\delta$ tends to have an effect on the model only close to a given input $x$ and that the large shifts towards other samples breaks the effect of $\delta$ on the model.

\whyinteresting

Defenses that employ stochastic elements are notoriously difficult to evaluate for two reasons. First, most attacks implicitly assume the targeted model to be deterministic. Breaking this assumption is a common cause for attack failure. Second, the quality of an adversarial perturbation now has to be measured as the average success rate, i.e. how often it can fool the model, which makes it difficult to select good adversarial perturbations (e.g. across repetitions and attacks). The Mixup defense is a particularly interesting instantiation of a stochastic defense because of its non-local mixing mechanism that interpolates between distant images.

\hypothesis

A large portion of the paper's main body is dedicated to an oblivious threat model in which the attacker is assumed to be unaware of the defense. This is not an interesting scenario: defending against a na\"ive attacker is not hard. Evaluating such a scenario is only interesting as a sanity check, but we recommend against presenting these results in the main body.

There is a brief discussion of adaptive attacks. For an adversarially trained model, the results are very similar to the oblivious attack, which raises a red flag: an adaptive attack should almost always perform substantially better than an oblivious attack that ignores the defense. This behavior suggests that the adaptive attack is not well suited for this defense. The exact algorithm of the adaptive attack is left unclear: the paper describes the adaptive attack as an average across multiple adversarial examples generated by PGD against the same sample. In the paper's appendix, however, the adaptive attack is described as a single PGD attack in which the gradient is estimated across multiple linear mixtures. The source code reveals that the first method was used, and that is problematic. Averaging across adversarial perturbations that have each been produced by an ineffective attack is unlikely to lead to much better adversarial examples. A stronger approach would be to average the gradient in each step of the attack.

Taken together, these observations suggest that a correctly adapted attack algorithm that takes the defense mechanism into account might be able to break the defense.

\attack
We perform our evaluation on CIFAR-10 using the code released by the authors. The threat model we use is $\ell_\infty$ attacks bounded by $\epsilon = 8/255$. We evaluate the paper's strongest base model (Interpolated adversarial training) which reaches 42.5\% accuracy on untargeted attacks without the defense, compared to 63.4\% (oblivious attack) or roughly 57\% (adaptive attack) for the defended model (MI-OL). We only evaluate the OL defense as suggested by the authors (private communication).

We include the mixing step into our attack. During the attack we backpropagate through the complete mixing mechanism using the same number of samples as used during evaluation. We applied PGD with 50 steps and a step-size of 1/255. During the attack and during evaluation of the perturbations we choose the mixing images randomly in every step. This procedure was sufficient to reduce the accuracy of the model to 43.9\%, very close to the accuracy of the base model without the Mixup defense. It is likely that one can further reduce accuracy through additional repetitions or steps, but due to the computational overhead of the Mixup mechanism, we did not perform additional experiments.

\lessons
\begin{myenumerate}
	\item The majority of the evaluation should focus on adaptive attacks, and not on attacks that are oblivious to the defense mechanism. Experiments with oblivious attacks are helpful sanity check and we recommend that authors include them in an appendix.
	\item For stochastic defenses, rather than averaging over the results of $n$ failed gradient-based attacks, it is better to stabilize the gradients of the model by averaging over $n$ queries. Only a stable gradient will allow an iterative gradient-based attack to succeed.
\end{myenumerate}

\exercise

\begin{myenumerate}
	\item Additional repetitions of PGD can typically reduce model accuracy by a few percentage points but for stochastic models care has to be taken as to which perturbation to choose. How much more can this model's accuracy be reduced using more attack repetitions?
	\item A standard component of attacks against stochastic defenses is EOT. Interestingly, we did not need to use EOT in this evaluation, but can it be used to further reduce model accuracy?
\end{myenumerate}

\section{ME-Net}
\label{sec:menet}

This paper~\citep{yang2019me} proposes a pre-processing step that randomly discards a large fraction of pixels in an image, and then uses matrix-estimation techniques to reconstruct the image. The defense trains a model on such pre-processed inputs, with the aim of learning representations that are less sensitive to small input variations.

\howitworks

Given an image $x$ represented as a matrix $M$, the defense first drops each entry in $M$ independently with some probability $p$ to obtain a noisy matrix $N$. It then reconstructs a matrix $\hat{M}$ from $N$ that should be close to $M$ in expectation.
Various matrix-estimation techniques are proposed. The ones we focus on here reconstruct $M$ using universal singular value thresholding (USVT)~\citep{chatterjee2015matrix} or nuclear-norm minimization~\citep{candes2009exact}. 

To train a model, $n$ random noisy matrices are generated for each training input $x$, and the respective matrix reconstructions $\hat{M}^{(1)}, \dots, \hat{M}^{(n)}$ are added to the training set. A standard classifier is then trained on this expanded dataset.
At inference, a single random noisy matrix is generated. The classifier processes the reconstruction of that matrix.

To combine the defense with adversarial training, the model is trained on adversarial examples from a PGD attack~\cite{madry2017towards} that uses the BPDA technique~\cite{athalye2018obfuscated} to backpropagate through the matrix estimation step. 
Here, instead of pre-generating $n$ matrix reconstructions for each training input, a noisy matrix and corresponding matrix reconstruction is computed for every step of PGD.

\whyinteresting

There have been a number of defenses proposed with a similar randomized pre-processing step~\citep{prakash2018deflecting, guo2018countering} which have been broken by adaptive attacks~\citep{athalye2018obfuscated, athalye2018cvpr}.
Yet, this paper does perform a thorough adaptive evaluation and concludes that the defense is sound.
 We thus viewed this defense as an interesting case-study: if we succeed in finding an attack on it, why did the adaptive attacks on similar prior defenses fail?

\hypothesis

Our first impression from this defense was simply that the proposed idea is too similar to previously broken defenses. At the same time, the paper does perform a fairly thorough evaluation of the defense, against a combination of black-box, white-box and adaptive attacks. Taking a closer look at this evaluation, we notice a few areas of concern. 

Since the defense's pre-processing is non-differentiable, the main white-box attack considered in the paper uses PGD with the BPDA technique of~\citep{athalye2018obfuscated}. To approximate gradients, the pre-processing is applied in the forward pass, but replaced by the identity function in the backward pass. The rationale for applying BPDA is that the pre-processing step (i.e., randomly dropping pixels and applying matrix estimation) computes a function that is close to the identity. While this use of BPDA appears appropriate, the paper does not take the most appropriate steps to deal with the defense's randomness. The defense is evaluated on attacks with multiple random restarts, but this is typically insufficient if the gradient computations at each step of the attack do not account for the defense's stochasticity.  In the same vein, the paper reports success in defending against
the score-based attack of~\citet{uesato2018adversarial} and the decision-based attack of~\citet{brendel2017decision}, both of which fare poorly against randomized defenses.

The paper considers two adaptive attacks. The first finds an adversarial perturbation for the reconstructed image, and then applies it to the original image. The second applies BPDA and then projects the gradient onto a low-rank subspace so that the attack focuses on global image structures. Both these attacks are found to perform significantly worse than PGD with BPDA, which demonstrates that they are not the most appropriate adaptive attacks.

We formulate the following hypotheses as to how this defense could be broken:

\begin{myenumerate}
	\item The attack should average over the randomness of the defense, using EOT (see \cref{sec:attack_background}).
	\item The paper seems to apply BPDA in a coarse fashion. As the initial randomized masking step is differentiable, a stronger attack might use BPDA to only approximate gradients of the matrix-estimation computation, rather than of the full pre-processing.
	\item Score-based attacks~\citep{ilyas2018black, chen2017zoo} might provide better gradients than BPDA, if combined with EOT.
\end{myenumerate}

We evaluate the defense on CIFAR-10, using code and models shared by the authors. 
We use the threat model from the paper, $\ell_\infty$ attacks with $\epsilon=8/255$. 
We focus our evaluation on the USVT approach, for which we could obtain a standard model as well as an adversarially trained model from the authors. We call these models USVT and USVT$_\text{ADV}$. After completing our evaluation, we further trained a standard model using nuclear-norm minimization and tested our best attack on that model as well.\footnote{Adversarial training with the nuclear-norm approach is extremely expensive (on the order of 10 CPU years) and we could not obtain a pre-trained model for this setting. So we refrain from evaluating it.}

We first reproduce the paper's white-box attack that uses BPDA to approximate the entire pre-processing step with the identity function. The approximated gradient is used in a PGD attack with $200$ steps. This attack reduces the models' accuracy to $35\%$ for USVT and $50\%$ for USVT$_\text{ADV}$, which is consistent with results reported in the paper.

We then attempted an attack that combines BPDA with EOT. We approximate the gradient $40$ times with different random masks, and use the average gradient in a PGD attack. This reduced the models' accuracies to $15\%$ for USVT and $32\%$ for USVT$_\text{ADV}$.
Note that the accuracy for USVT$_\text{ADV}$ is lower than that obtained with adversarial training alone (the same ResNet-18 model architecture achieves over $40\%$ robust accuracy with adversarial training). The reason for this is probably because the attack used during training (PGD+BPDA) is not strong enough to approximate worst-case adversarial examples for this defense.

\attack

We were curious as to why the BPDA+EOT attack did not have an even higher success rate for the standard model trained with USVT matrix estimation.

First, we found that the defense's randomness causes significant fluctuations in the model's outputs: when classifying inputs multiple times in a row, we find that there's a non-negligible chance of the model misclassifying an input due to a ``bad'' choice of random mask. This brittleness of the defense may also make it harder to produce robust adversarial examples.

We further investigated whether the assumption underlying the use of BPDA was valid, namely that the pre-processing step approximates the identity function. Surprisingly, we found that this is not necessarily the case for the USVT approach. For CIFAR-10 images, USVT fails to reconstruct an image close to the original. In fact, we find that the output of the pre-processing step is roughly identical to the randomly masked image. We thus tried to apply BPDA only to the USVT procedure, and backpropagate through the random mask. However, this attack did not improve upon the $15\%$ accuracy we reached when applying BPDA to the full pre-processing step (in fact, applying BPDA solely to USVT resulted in a slightly worse attack).

Taking a step back (and while trying to find bugs in our own attack implementation), we attempted to attack solely the trained classifier (a ResNet-18), without the pre-processing stage. Surprisingly, we found the base classifier to be moderately robust: For $\ell_\infty$-noise bounded by $\epsilon=8/255$, we could not get its accuracy below $10\%$. We further checked the model's gradients, and found them to be highly interpretable, a phenomenon previously demonstrated for adversarially trained models~\citep{tsipras2018robustness}. The non-trivial robustness of this base classifier is quite remarkable, as it was trained in a standard fashion on randomly masked and reconstructed images. This gives credence to the authors' intuition that training on randomly masked and reconstructed images can encourage a model to learn somewhat more robust representations. 
Altogether, this suggests that an attack on the full defense is unlikely to bring the model's accuracy much further down than our best attack.

Finally, we evaluated a model trained with nuclear-norm minimization. Due to the high computational cost of that defense (the nuclear-norm minimization is CPU-bound and takes about one second per input), we refrained from performing any additional adaptive experiments, and simply used the best attack we found for the USVT model (i.e., BPDA+EOT). This attack reduced the accuracy of the nuclear-norm defense to $13\%$, a similar result as for the defense instantiated with USVT. 

\lessons

\begin{myenumerate}
	\item Attacks must account for all of the randomness of a defense.
	\item For defenses that train a base classifier with additional components, an ablation study can reveal where the increased robustness comes from.
\end{myenumerate}

\exercise

\begin{myenumerate}
	\item Why does training on masked
and reconstructed inputs led to a base classifier with low yet non-trivial robustness?
Or can the base classifier be attacked with a stronger attack?

	\item Is there a differentiable version of USVT so as to remove the need for BPDA in the attack? 
\end{myenumerate}

\section{Asymmetrical Adversarial Training}
\label{sec:asymmetrical}

This paper \citep{yin2020adversarial} proposes to use adversarially-trained models to detect adversarial examples.
The idea is a natural extension to the work of~\citet{madry2017towards}. 

\howitworks

For a $K$-class problem, the defense uses $K$ detector models $h_1, \dots, h_K$. Given an input $x$, the $i$-th detector outputs a logit score $h_i(x) \in \mathbb{R}$ for class $i$. 
Let $\sigma(\cdot)$ denote the sigmoid function.
The detectors are adversarially trained so that for each training input $(x, y)$, we maximize
$\sigma(h_{y}(x))$ and minimize the terms $\max_{||\delta||\leq \epsilon} \sigma(h_{i}(x + \delta))$, for all $i \neq y$.
The first term ensures that the detector of the correct class $y$ recognizes $x$ as benign. The other terms encourage the detectors of classes $i \neq y$ to reject perturbed versions of $x$.

The paper proposes two classification frameworks:
\begin{myenumerate}
	\item The \emph{integrated classifier} uses a standard ``base'' classifier $f(x)$ to output a prediction $f(x)=y$. The input is then fed to the detector for that class, $h_y$, and rejected if the logit $h_y(x)$ is below some threshold $\tau$.
	\item The \emph{generative classifier} computes a logit vector from the scores of all detectors. The classifier's prediction is $f(x) = \argmax_i\ h_i(x)$ and low confidence inputs are rejected (i.e., if $\max h_i(x) < \tau$ for some threshold $\tau$).
\end{myenumerate}

\whyinteresting
This defense is interesting in that the authors seemed to have done everything right. The proposed adaptive attack (see below) is well motivated and introduces no unnecessary complexity. 
We initially believed that this defense would resist our own adaptive attacks.

\hypothesis

Compared to many other defenses, this ensemble of adversarially-trained detectors is fairly simple (a positive trait in our opinion). The classifier is built from $K$ standard models, none of which should (a priori) show any signs of gradient masking or other optimization difficulties. The defense's evaluation is also quite appropriate. The detectors are first evaluated in isolation using standard attacks, and the authors further propose an adaptive attack with a loss function tailored to the full defense.
Finally, the authors show that their robust detectors are \emph{invertible}. That is, an unbounded attack that maximizes the score of a detector results in semantically meaningful images from that class---mirroring similar results for adversarially trained classifiers~\citep{santurkar2019computer}. Overall, this gives good credence that this defense is indeed robust. 

A closer look at the defense's evaluation does however reveal one issue: a sub-optimal choice of loss function in the considered adaptive attack.
Let $\logits(x)_i \in \mathbb{R}$ denote the $i$-th logit of the base classifier.
Given an input $(x, y)$, the attack on the integrated classifier finds an adversarial example $x'$ that maximizes:
\[
L(x', y) = 
\begin{cases}
\max\limits_{i \neq y}\logits(x')_i - \logits(x')_y & \text{if } f(x') = y \\
\max\limits_{i \neq y} h_{i}(x') & \text{otherwise} 
\end{cases}\;.
\]
In words, the attack first finds an $x'$ that is misclassified by the base classifier, using a standard loss function proposed in~\citep{carlini2017towards}. Then, once the base classifier is fooled, the attack further maximizes the scores of all detectors other than that of the true class $y$, to ensure that the example is not rejected.

The issue with this loss function is that the second term, $\max_{i \neq y} h_{i}(x')$, does not capture the true attack objective. Indeed, to bypass the defense, we only need to fool the detector of the class predicted by the base classifier, i.e., $y' = f(x')$. Maximizing the scores of all detectors is thus wasteful of the attack's limited perturbation budget.

The adaptive attack on the generative classifier has a similar issue. Here, the proposed loss function is simply:
\[
L(x', y) = \max\limits_{i \neq y} h_{i}(x') \;.
\]
That is, the attack maximizes the score of the most confident detector other than the one of the true class. The issue is that even with an unbounded perturbation, maximizing this loss may fail to produce a successful attack.
For example, one way to maximize this loss function is to increase each detectors' score by the same value, thereby leaving the classifier's prediction unchanged.

\attack

We evaluate pre-trained CIFAR-10 detectors released by the authors. We report results for  $\ell_\infty$ attacks with $\epsilon=8/255$.

We first describe our attack on the integrated classifier. A natural idea for improving the loss function from the paper would be to first fool the base classifier into outputting some class $f(x') \neq y$, and then to maximize the score of the detector for class $f(x')$:
\[
L(x', y) = 
\begin{cases}
\max\limits_{i \neq y} \logits(x')_{i} - \logits(x')_y & \text{if } f(x') = y \\
h_{f(x')}(x') & \text{otherwise} 
\end{cases}\;.
\]
However, we tend to dislike loss functions of this type as work is still wasted if the classifier's prediction $f(x')$ ever switches from some class $i\neq y$ to another class $j \neq y$.
Instead, we formulate a \emph{targeted} attack objective:
\begin{align*}
L_{t}(x') &= \logits(x')_t - \max\limits_{i \neq t} \logits(x')_{i} \\
L(x', t) &= 
\begin{cases}
 L_{t}(x') & \text{if } L_{t}(x') < \kappa \\
 h_{t}(x') & \text{otherwise} 
\end{cases}\;.
\end{align*}
That is, we first fool the base classifier into outputting the target class $t$, and then maximize the score of the detector for the target class. 
The small constant $\kappa>0$ ensures that $f$'s targeted misclassification is not overly brittle.
We stop the optimization on the base classifier $f$ once a targeted misclassification is achieved since the confidence of $f$ is not considered by the defense. 

We then perform a \emph{multi-targeted} attack~\citep{gowal2019alternative, carlini2019ami}: for each target class $t \neq y$, we maximize $L(x', t)$ using $100$ PGD steps, and we retain the best adversarial example across all $K-1$ attacks.

This attack is substantially more effective than the one in the paper's evaluation.
At a False-Positive-Rate of $5\%$, we reduce the detector's robust accuracy from $30\%$ to $11\%$. 

For the generative classifier, we also used a multi-targeted attack as it is typically very effective~\citep{gowal2019alternative}. Our objective maximizes the score of the targeted detector, while minimizing the highest detector score, until the targeted detector's score is highest. At that point, we further maximize the target detector's score to ensure a confident prediction:
 \begin{align*}
  L_t(x') &= 
 h_t(x') - \max\limits_{i \neq t} h_i(x')
\\ 
 L(x', t) &= 
 \begin{cases}
 L_t(x') & \text{if } L_t(x') < \kappa \\
 h_{t}(x') & \text{otherwise} 
 \end{cases}\;.
 \end{align*}
At a False-Positive-Rate of $5\%$, the multi-targeted attack (with 100 steps of PGD per target) reduces the generative classifier's robust accuracy from $55\%$ to $37\%$, lower than that of full PGD adversarial training. 

This suggests that the attacks used at training time might not be optimal either. Recall that the defense is trained so as to maximize $\sigma(h_{y}(x))$ and minimize $\max_{||\delta||\leq \epsilon} \sigma(h_{i}(x + \delta))$ for $i \neq y$. Intuitively, while this ensures that the detectors for class $i \neq y$ have low confidence on adversarial examples, it does not ensure that the detector for the true class has high confidence on adversarial examples. Adversarially training \emph{all} detectors for each input might lead to a more robust defense.

Given that our attacks did not fully reduce the accuracy of either model to $0\%$, one may wonder why 
we are confident that our evaluation is sufficiently thorough that we have accurately
evaluated the defense's robustness to within a small margin. 
As this defense is not certified, our evaluation will ultimately always be empirical. 
We believe it unlikely we can significantly reduce the defense's accuracy further due to the following:
\begin{myitemize}
	\item The defense's simplicity, which does not suggest any gradient masking issues.
	\item The defense's strong interpretability---as demonstrated by the authors---which had been previously observed in other robust models.
	\item Our attacks' simplicity, which reduces the risk of implementation errors or of wasting perturbation budget on unnecessary objectives.
	\item The defenses' resilience under standard checks~\citep{carlini2019evaluating}:  increasing the number of PGD iterations (to $200$) and random restarts (to $10$), only reduce the defense's accuracy by an extra $1\%$.
\end{myitemize}

\lessons

\begin{myenumerate}
	\item A good loss function has the property that increasing the loss always increases the attack success rate. Loss functions without this property can ``waste'' the perturbation budget on orthogonal objectives.
	\item It is sometimes easier to write down a good targeted loss function than an untargeted one. A ``multi-targeted'' attack that targets each class in turn will be more effective than an untargeted attack in such a case.
\end{myenumerate}

\exercise

\begin{myenumerate}
	\item Can the accuracy of the defense be reduced further, especially for the generative classifier?
	\item Are there other defenses that can be attacked by formulating an untargeted attack as a series of multiple targeted attacks?
\end{myenumerate}

\section{Turning a Weakness into a Strength}
\label{sec:weakness}

This defense~\citep{hu2019new} is an adversarial example detector.
While the defense is conceptually simple, the original evaluation methodology is
complex and constructing a single unified loss function that encodes the attacker's objective is challenging.

\howitworks

Note that in this section, we sometimes use $f(x)$ to refer to either the classifier's output class, or the full vector of class probabilities. The choice should be clear from context. We denote by $f(x)_i$ the probability that the classifier assigns to class $i$.
To determine if a given input $x$ is adversarial, the defense sets thresholds $\tau_1$ and $\tau_2$ and
checks if either of the following are true:
\begin{enumerate}
\item $\sum_i \lVert f(x + \delta)_i - f(x)_i \rVert_1 > \tau_1$ for $\delta \sim \mathcal{N}(0, I \sigma^2)$; or,
\item $\texttt{\#Steps}(f, x) > \tau_2$ where $\texttt{\#Steps}$ returns
  the number of steps required to generate an adversarial example
  $x'$ such that either (a) $f(x) \ne y$ or (b) $f(x) = t$.
\end{enumerate}

The first term is simple: the models predictions should not change
substantially when the input is slightly perturbed by Gaussian noise.
An idea related to this term alone has previously been suggested 
as a defense \citep{cao2017mitigating} and was subsequently broken \citep{he2018decision}.

The second loss term is more complicated: given an input $x$ we compute
the number of steps necessary to turn $x$ into an adversarial example $x'$.
Intuitively, this check enforces that $x$ is \emph{not} high confidence:
if it is too far from the decision boundary, then this test will fail.
This would have prevented He \emph{et al.}'s attack.

\whyinteresting
This defense is conceptually simple, but by construction it is hard
to define a straightforward differentiable loss function that can
be minimized to generate adversarial examples against it.
This defense relies on ideas similar to the defense in \cref{sec:odds}.

\hypothesis
The paper contains an adaptive attack by developing a new loss function
that is designed to evade the defense.
This loss function is defined as a combination of four independent
loss terms
$L = \lambda L_1 + L_2 + L_3 + L_4$.

The first term $L_1$ is a standard cross-entropy loss to fool the classifier,
$L_1 = L_{CE}(f(x), y_{\text{true}})$.
The second loss term encourages the classifier to be confident in its prediction
by directly optimizing the first detection term
$L_2 = \mathbb{E}_{\delta \sim \mathcal{N}(0, I \sigma^2)} \left
[ \lVert f(x) - f(x' + \delta) \rVert_1 \right ]$ in order to defeat
the first detector.
The third loss term encourages the classifier to have a high loss
after taking a single gradient descent step of size $\alpha$
towards any incorrect class $y'$
$L_3 = -\mathbb{E}_{y' \ne y_t} L_{CE}(f(x - \alpha \nabla_x L_{CE}(f(x), y')), y_t)$
and the final loss term encourages the same but for the target
class $y_t$.
$L_4 = -L_{CE}(f(x + \alpha \nabla_x L_{CE}(f(x), y_t)), y_t)$
Loss functions $L_3$ and $L_4$ are non-differentiable, and so the paper
applies BPDA \citep{athalye2018obfuscated}.

This combined loss function is exceptionally complicated, and it is unlikely that
optimizing it would be able to minimize all four terms simultaneously.
The lack of additional weighting terms $\lambda$ on each hyper-parameter
made us suspect that when generating an adversarial examples,
$L_1$ would dominate and the remaining loss terms would not actually
change the final resulting adversarial example.
Combining just two terms together can often be difficult \citep{carlini2017adversarial} and
requires adaptively tuning the hyper-parameter $\lambda$; combining
four introduces even more complexity.

The paper applies PGD using the
standard cross-entropy loss \citep{madry2017towards} and the logit margin loss \citep{carlini2017towards}.
Typically, the difference between these two losses is small: a few
percentage points or less.
However this paper finds a large difference in many settings.
For example, at a distortion bound of $\epsilon=0.1$ the PGD on
the cross entropy loss succeeds just $8.5\%$ of the time.
In comparison, the margin loss succeeds over $32\%$ of the time.
This large of a gap is atypical; we suspected the root cause to be 
a non-standard
experimental methodology or attack implementation.

As a result, we inspected the open-source implementation
provided by the authors.
The paper implements its own attack algorithms instead of calling a
primitive from some reference attack library such as Foolbox~\citep{rauber2017foolbox} or CleverHans~\citep{papernot2016technical}.
We find that the implemented attack does not take the sign of the gradient
when implementing $\ell_\infty$ PGD \citep{madry2017towards}.
The typical definition of $\ell_\infty$ PGD is to define 
$\text{normalize}(x) = \text{sign}(x)$.
The implementation of PGD contained in this paper does not include
the sign operator and does \emph{not} normalize the gradient.
Importantly, the paper makes no definition of PGD and as such does
not explicitly state if their PGD method includes this normalization step.%
\footnote{In personal communication with the authors they state
  that they intended for PGD to \emph{not} have the sign operator.}

Thus, this explains the artifact we observed where
the logit margin loss is
superior to cross-entropy loss.
The magnitude of the gradient of the cross entropy loss can vary
by several orders of magnitude.
When the classifier is confident in its prediction, as is typically
the case for poorly-calibrated models, $\lVert \nabla L_{CE} \rVert$
is often exceptionally small: $10^{-5}$ or lower.
In contrast, when the classifier is not confident and the top two
predictions are almost equally likely, $\lVert \nabla L _{CE}\rVert$
can be as large as $1$.
The gradient of the margin loss 
is much better behaved: it tends to not change by more than an
order of magnitude.

More concerning than the fact that the implementation of PGD
used to attack the network does not contain the sign operator,
the implementation of PGD used in phase (2) \emph{of the defense}
(that computes $\text{\#Steps}$)
omits the sign operator.
Because the defense uses the number of steps of gradient descent
as a proxy for the $\ell_p$ distance to the decision boundary itself
(instead of using the distance to the decision boundary directly), this implementation difference
is extremely important.
The defense would probably have been more effective if it had used the standard
attack implementation with the sign on the gradient.

\attack

The core assumption the defense relies on is that there do not exist
inputs that
(a) have high confidence on random noise, but (b) are still close to
the decision boundary when running an adversarial attack.
One option would be to mount a feature adversary attack as we did against 
a similar defense in~\cref{sec:odds}.
However, in this section we aim to mount a new type of adaptive attack.
We set out to directly construct an input with properties (a) and (b) above, 
by initially
generating an adversarial example with high confidence and then slowly
moving it towards the decision boundary.

For an input example $x$ we initially generate a high-confidence
adversarial example $x'$ by running $100$ steps of PGD (\emph{with} the sign
on the gradient).
Then,
we let $x_\alpha = \alpha \cdot x' + (1-\alpha) \cdot x$ linearly interpolate
between the original example and the adversarial example,
such that $x_0 = x$ and $x_1 = x'$.
We find that often, there exists a choice $x_{\alpha^*}$ that it is
still high confidence but also is sufficiently close to the decision boundary
that the defense will not reject it.

Our first attack simply samples $1{,}000$ values of $\alpha$ between
$0$ and $1$ and checks if the defense's checks (1) and (2) are both satisfied for
any of these examples.
There are two problems with this approach.
First, it is computationally expensive: the detection method has an inline loop
that runs PGD and therefore to generate a single
adversarial example requires several minutes.
Worse, we find that it sometimes fails when a more fine-grained search
would have succeeded, because the necessary value of $\alpha$ 
to fool the defense is very small, i.e., $\alpha < 10^{-4}$.
Increasing the number of steps is one possible solution---which we used
to diagnose the failure here---but is even slower.

To improve the computational complexity, we instead
perform a binary search to find a value of $\alpha$
so that both defense properties (1) and (2) hold.
We initialize $\alpha_{\text{low}}=0$ and $\alpha_{\text{high}}=1$.
We define $\alpha_{\text{mid}}=(\alpha_{\text{low}} + \alpha_{\text{high}})/2$
and evaluate $x_{\alpha_{\text{mid}}}$.
If property (1) holds but property (2) does not, then we know that
the current input is robust to noise but too far away from the decision
boundary, and thus we would like to move it closer to the boundary.
So we set $\alpha_{\text{low}}=\alpha_{\text{mid}}$ and recurse.
Conversely, if property (1) does not hold but property (2) does hold,
then we are too close to the decision boundary and need to back away.
We set $\alpha_{\text{high}}=\alpha_{\text{mid}}$ and recurse.
If both properties hold, we have a successful adversarial example.
If neither hold, we randomly generate a new adversarial example $x'$ and repeat
the binary search procedure for this example.

The most important reason why this attack is effective is that it is
simple to analyze.
It can be difficult to diagnose why gradient descent fails,
and so designing attacks that are as simple as possible makes it
easy to diagnose why they are not working effectively.
When simple techniques do not work as expected, it is easy
to learn the correct lesson for why they failed.
In contrast, when complex techniques do not work as
expected, one has to consider multiple hypotheses and evaluate
each one to understand the true reason for failure.

This attack is successful at reducing the accuracy of the classifier
on CIFAR-10 to $0\%$ at a $0\%$ detection rate, and similarly 
bring the ImageNet classifier to $<1\%$ accuracy at a $0\%$ detection
rate, both in the threat model originally considered by the paper.

\lessons

\begin{myenumerate}
\item Combining multiple terms together leads to difficult-to-optimize
  loss functions that may not behave as desired.

\item BPDA should not be treated as a general-purpose method for minimizing
  through arbitrarily non-differentiable functions.
  Rather, loss functions must be designed to work with BPDA and must
  already be \emph{almost differentiable}.
\end{myenumerate}

\exercise

\begin{myenumerate}
\item Can the proposed defense be defeated with a feature adversary, instead of the special-purpose attack we construct?
\item If the defense had applied standard PGD, taking the sign
of the gradient before gradient steps, would it be more robust?
\end{myenumerate}

\section{Conclusion}

We see a marked improvement in defense evaluations compared to those
studied in prior work \citep{carlini2017adversarial,athalye2018obfuscated}:
whereas in the past, papers would simply not perform an evaluation with a strong
adaptive attack,
nearly all defense evaluations we study \emph{do} perform adaptive attacks
that aim to evade the proposed defense.
However, 
we find that despite the \emph{presence} of an adaptive attack evaluation, all
\numdefenses defenses we analyzed could be circumvented
with improved attacks.

We have described a number of attack strategies that proved useful in circumventing different defenses. Yet, we urge the community to refrain from using these  adaptive attacks as templates to be copied in future
evaluations. To illustrate why, we propose the following informal ``no-free-lunch-theorem'':
\begingroup
\addtolength\leftmargini{-0.3in}
\begin{quote}
	\emph{For any proposed attack, it is possible to build a non-robust defense that prevents that attack.}
\end{quote}
\endgroup
Such a defense is generally not interesting. We thus encourage viewing robustness evaluations against prior attacks (including adaptive attacks on prior defenses) only as a useful sanity check.
The main focus of a robustness evaluation should be on developing comprehensive adaptive attacks that explicitly uncover and target the defense's weakest links (by making use of prior techniques only if appropriate).
In particular, we strongly encourage authors to focus their evaluation \textbf{solely} on adaptive attacks (and defer evaluations based on non-adaptive attacks to an appendix).
It is nevertheless important to include a non-adaptive evaluation to demonstrate that simple methods are insufficient.
We hope that this paper can serve as a guide to
performing and evaluating such adaptive attacks.

It is critical that adaptive attacks are hand-designed to target specific
defenses.
No automated tool will be able to comprehensively evaluate the robustness
of a defense, no matter how sophisticated.
For example, the state-of-the-art in automated attacks,
``AutoAttack'' \citep{croce2020reliable}, evaluates four of the same defenses we do. In two cases, AutoAttack only partially succeeds in attacking the proposed defenses when we break them completely. Furthermore, existing automated tools like AutoAttack cannot be directly applied to many of the defenses we studied, such as those that detect adversarial examples.

Perhaps the most important lesson from our analysis is that 
adaptive attacks should be as \emph{simple} as possible,
while resolving any potential optimization difficulties.
Each new component added on top of straightforward gradient descent
is another place where errors can arise. 
On the whole, the progress in defense evaluations
over the past years is encouraging, and we hope that stronger
adaptive attacks evaluations will pave the way to more robust models.

\section*{Broader Impact}

Research that studies the security of machine learning, and especially papers
whose primary purpose is to develop attacks on proposed defenses, must be
careful to do more good than harm.
We believe that this will be the case with our paper.
After decades of debate, the computer security community has largely converged
on ``responsible disclosure'' as the optimal method for disclosing vulnerabilities:
after discovering a vulnerability, responsible disclosure dictates that
the affected parties should be notified fist, and after a reasonable
amount of time, the disclosure should be made public so that the
community as a whole can learn from it.

We notified all authors of our breaks of their defenses before making our paper
public.
Authors from twelve of the thirteen papers responded to us and verified that
our evaluations were accurate (we offered to provide the generated adversarial
examples to all authors).
Further, we do not believe that there are any deployed systems that rely
on the security of any of these particular defenses.

However, it remains a possibility that our methodology for constructing attacks
could be used to break some other system which has been deployed, or will be deployed in the future.
This is unavoidable, however we firmly believe that the help that our paper can provide to researchers designing new defenses significantly outweighs the help that it may provide an actual malicious actor.
Our paper is focused on assisting researchers perform more thorough evaluations,
and diagnosing failures in evaluations---not on attacking real systems or users.

\section*{Acknowledgments}

We thank the authors of all the defenses we studied for helpful discussions, for sharing code and pre-trained models, and for comments on early drafts of this paper, and the anonymous reviewers for their comments.
  
Florian Tram\`er's research was supported in part by the Swiss National Science Foundation (SNSF project P1SKP2\_178149), NSF award CNS-1804222 and a gift from the Open Philanthropy Project. Nicholas Carlini's research is supported by Google. Wieland Brendel's research was supported by the German Federal Ministry of Education and Research through the T\"ubingen AI Center (FKZ 01IS18039A) as well as by the Intelligence Advanced Research Projects Activity (IARPA) via Department of Interior/Interior Business Center (DoI/IBC) contract number D16PC00003. The U.S. Government is authorized to reproduce and distribute reprints for Governmental purposes notwithstanding any copyright annotation thereon. Disclaimer: The views and conclusions contained herein are those of the authors and should not be interpreted as necessarily representing the official policies or endorsements, either expressed or implied, of IARPA, DoI/IBC, or the U.S. Government. Aleksander M\k{a}dry's research was supported in part by the NSF awards CCF-1553428 and CNS-1815221. This material is based upon work supported by the Defense Advanced Research Projects Agency (DARPA) under Contract No. HR001120C0015.

\bibliography{biblio}

\begin{thebibliography}{59}
\providecommand{\natexlab}[1]{#1}
\providecommand{\url}[1]{\texttt{#1}}
\expandafter\ifx\csname urlstyle\endcsname\relax
  \providecommand{\doi}[1]{doi: #1}\else
  \providecommand{\doi}{doi: \begingroup \urlstyle{rm}\Url}\fi

\bibitem[Athalye \& Carlini(2018)Athalye and Carlini]{athalye2018cvpr}
Athalye, A. and Carlini, N.
\newblock On the robustness of the cvpr 2018 white-box adversarial example
  defenses.
\newblock In \emph{Computer Vision: Challenges and Opportunities for Privacy
  and Security,}, 2018.

\bibitem[Athalye et~al.(2018{\natexlab{a}})Athalye, Carlini, and
  Wagner]{athalye2018obfuscated}
Athalye, A., Carlini, N., and Wagner, D.
\newblock Obfuscated gradients give a false sense of security: Circumventing
  defenses to adversarial examples.
\newblock In \emph{International Conference on Machine Learning},
  2018{\natexlab{a}}.

\bibitem[Athalye et~al.(2018{\natexlab{b}})Athalye, Engstrom, Ilyas, and
  Kwok]{athalye2017synthesizing}
Athalye, A., Engstrom, L., Ilyas, A., and Kwok, K.
\newblock Synthesizing robust adversarial examples.
\newblock In \emph{International Conference on Machine Learning},
  2018{\natexlab{b}}.

\bibitem[Bafna et~al.(2018)Bafna, Murtagh, and Vyas]{bafna2018thwarting}
Bafna, M., Murtagh, J., and Vyas, N.
\newblock Thwarting adversarial examples: An l0-robust sparse fourier
  transform.
\newblock In \emph{Advances in Neural Information Processing Systems}, pp.\
  10096--10106, 2018.

\bibitem[Biggio et~al.(2013)Biggio, Corona, Maiorca, Nelson, {\v{S}}rndi{\'c},
  Laskov, Giacinto, and Roli]{biggio2013evasion}
Biggio, B., Corona, I., Maiorca, D., Nelson, B., {\v{S}}rndi{\'c}, N., Laskov,
  P., Giacinto, G., and Roli, F.
\newblock Evasion attacks against machine learning at test time.
\newblock In \emph{Joint European conference on machine learning and knowledge
  discovery in databases}, pp.\  387--402. Springer, 2013.

\bibitem[Brendel et~al.(2018)Brendel, Rauber, and Bethge]{brendel2017decision}
Brendel, W., Rauber, J., and Bethge, M.
\newblock Decision-based adversarial attacks: Reliable attacks against
  black-box machine learning models.
\newblock \emph{International Conference on Learning Representations}, 2018.

\bibitem[Brendel et~al.(2019)Brendel, Rauber, Matthias, Ustyuzhaninov, and
  Bethge]{brendel2019bethge}
Brendel, W., Rauber, J., Matthias, K., Ustyuzhaninov, I., and Bethge, M.
\newblock Accurate, reliable and fast robustness evaluation.
\newblock \emph{33rd Conference on Neural Information Processing Systems
  (NeurIPS)}, 2019.

\bibitem[Cand{\`e}s \& Recht(2009)Cand{\`e}s and Recht]{candes2009exact}
Cand{\`e}s, E.~J. and Recht, B.
\newblock Exact matrix completion via convex optimization.
\newblock \emph{Foundations of Computational mathematics}, 9\penalty0
  (6):\penalty0 717, 2009.

\bibitem[Cao \& Gong(2017)Cao and Gong]{cao2017mitigating}
Cao, X. and Gong, N.~Z.
\newblock Mitigating evasion attacks to deep neural networks via region-based
  classification.
\newblock In \emph{Proceedings of the 33rd Annual Computer Security
  Applications Conference}, pp.\  278--287, 2017.

\bibitem[Carlini(2019)]{carlini2019ami}
Carlini, N.
\newblock Is {AmI} (attacks meet interpretability) robust to adversarial
  examples?
\newblock \emph{arXiv preprint arXiv:1902.02322}, 2019.

\bibitem[Carlini \& Wagner(2016)Carlini and Wagner]{carlini2016defensive}
Carlini, N. and Wagner, D.
\newblock Defensive distillation is not robust to adversarial examples.
\newblock \emph{arXiv preprint arXiv:1607.04311}, 2016.

\bibitem[Carlini \& Wagner(2017{\natexlab{a}})Carlini and
  Wagner]{carlini2017adversarial}
Carlini, N. and Wagner, D.
\newblock Adversarial examples are not easily detected: Bypassing ten detection
  methods.
\newblock In \emph{Proceedings of the 10th ACM Workshop on Artificial
  Intelligence and Security}, pp.\  3--14, 2017{\natexlab{a}}.

\bibitem[Carlini \& Wagner(2017{\natexlab{b}})Carlini and
  Wagner]{carlini2017towards}
Carlini, N. and Wagner, D.
\newblock Towards evaluating the robustness of neural networks.
\newblock In \emph{2017 IEEE symposium on security and privacy}, pp.\  39--57.
  IEEE, 2017{\natexlab{b}}.

\bibitem[Carlini et~al.(2019)Carlini, Athalye, Papernot, Brendel, Rauber,
  Tsipras, Goodfellow, and Madry]{carlini2019evaluating}
Carlini, N., Athalye, A., Papernot, N., Brendel, W., Rauber, J., Tsipras, D.,
  Goodfellow, I., and Madry, A.
\newblock On evaluating adversarial robustness.
\newblock \emph{arXiv preprint arXiv:1902.06705}, 2019.

\bibitem[Chatterjee et~al.(2015)]{chatterjee2015matrix}
Chatterjee, S. et~al.
\newblock Matrix estimation by universal singular value thresholding.
\newblock \emph{The Annals of Statistics}, 43\penalty0 (1):\penalty0 177--214,
  2015.

\bibitem[Chen et~al.(2017)Chen, Zhang, Sharma, Yi, and Hsieh]{chen2017zoo}
Chen, P.-Y., Zhang, H., Sharma, Y., Yi, J., and Hsieh, C.-J.
\newblock Zoo: Zeroth order optimization based black-box attacks to deep neural
  networks without training substitute models.
\newblock In \emph{Proceedings of the 10th ACM Workshop on Artificial
  Intelligence and Security}, pp.\  15--26, 2017.

\bibitem[Chen et~al.(2018)Chen, Sharma, Zhang, Yi, and Hsieh]{chen2018ead}
Chen, P.-Y., Sharma, Y., Zhang, H., Yi, J., and Hsieh, C.-J.
\newblock Ead: elastic-net attacks to deep neural networks via adversarial
  examples.
\newblock In \emph{Thirty-second AAAI conference on artificial intelligence},
  2018.

\bibitem[Croce \& Hein(2020)Croce and Hein]{croce2020reliable}
Croce, F. and Hein, M.
\newblock Reliable evaluation of adversarial robustness with an ensemble of
  diverse parameter-free attacks.
\newblock In \emph{ICML}, 2020.

\bibitem[Dhillon et~al.(2018)Dhillon, Azizzadenesheli, Lipton, Bernstein,
  Kossaifi, Khanna, and Anandkumar]{dhillon2018stochastic}
Dhillon, G.~S., Azizzadenesheli, K., Lipton, Z.~C., Bernstein, J.~D., Kossaifi,
  J., Khanna, A., and Anandkumar, A.
\newblock Stochastic activation pruning for robust adversarial defense.
\newblock In \emph{International Conference on Learning Representations}, 2018.

\bibitem[Dong et~al.(2018)Dong, Liao, Pang, Su, Zhu, Hu, and
  Li]{dong2018boosting}
Dong, Y., Liao, F., Pang, T., Su, H., Zhu, J., Hu, X., and Li, J.
\newblock Boosting adversarial attacks with momentum.
\newblock In \emph{Proceedings of the IEEE conference on computer vision and
  pattern recognition}, pp.\  9185--9193, 2018.

\bibitem[Engstrom et~al.(2019)Engstrom, Ilyas, Santurkar, Tsipras, Tran, and
  Madry]{engstrom2019adversarial}
Engstrom, L., Ilyas, A., Santurkar, S., Tsipras, D., Tran, B., and Madry, A.
\newblock Adversarial robustness as a prior for learned representations.
\newblock \emph{arXiv preprint arXiv:1906.00945}, 2019.

\bibitem[Gowal et~al.(2019)Gowal, Uesato, Qin, Huang, Mann, and
  Kohli]{gowal2019alternative}
Gowal, S., Uesato, J., Qin, C., Huang, P.-S., Mann, T., and Kohli, P.
\newblock An alternative surrogate loss for {PGD}-based adversarial testing.
\newblock \emph{arXiv preprint arXiv:1910.09338}, 2019.

\bibitem[Guo et~al.(2018)Guo, Rana, Cisse, and van~der
  Maaten]{guo2018countering}
Guo, C., Rana, M., Cisse, M., and van~der Maaten, L.
\newblock Countering adversarial images using input transformations.
\newblock In \emph{International Conference on Learning Representations}, 2018.

\bibitem[He et~al.(2017)He, Wei, Chen, Carlini, and Song]{he2017adversarial}
He, W., Wei, J., Chen, X., Carlini, N., and Song, D.
\newblock Adversarial example defense: Ensembles of weak defenses are not
  strong.
\newblock In \emph{11th $\{$USENIX$\}$ Workshop on Offensive Technologies
  ($\{$WOOT$\}$ 17)}, 2017.

\bibitem[He et~al.(2018)He, Li, and Song]{he2018decision}
He, W., Li, B., and Song, D.
\newblock Decision boundary analysis of adversarial examples.
\newblock In \emph{International Conference on Learning Representations}, 2018.

\bibitem[Hosseini et~al.(2019)Hosseini, Kannan, and
  Poovendran]{hosseini2019odds}
Hosseini, H., Kannan, S., and Poovendran, R.
\newblock Are odds really odd? bypassing statistical detection of adversarial
  examples.
\newblock \emph{arXiv preprint arXiv:1907.12138}, 2019.

\bibitem[Hu et~al.(2019)Hu, Yu, Guo, Chao, and Weinberger]{hu2019new}
Hu, S., Yu, T., Guo, C., Chao, W.-L., and Weinberger, K.~Q.
\newblock A new defense against adversarial images: Turning a weakness into a
  strength.
\newblock In \emph{Advances in Neural Information Processing Systems}, pp.\
  1633--1644, 2019.

\bibitem[Ilyas et~al.(2018)Ilyas, Engstrom, Athalye, and Lin]{ilyas2018black}
Ilyas, A., Engstrom, L., Athalye, A., and Lin, J.
\newblock Black-box adversarial attacks with limited queries and information.
\newblock In \emph{International Conference on Machine Learning}, pp.\
  2137--2146, 2018.

\bibitem[Jacobsen et~al.(2019)Jacobsen, Behrmannn, Carlini, Tramer, and
  Papernot]{jacobsen2019exploiting}
Jacobsen, J.-H., Behrmannn, J., Carlini, N., Tramer, F., and Papernot, N.
\newblock Exploiting excessive invariance caused by norm-bounded adversarial
  robustness.
\newblock \emph{arXiv preprint arXiv:1903.10484}, 2019.

\bibitem[Kingma \& Welling(2014)Kingma and Welling]{kingma2013auto}
Kingma, D.~P. and Welling, M.
\newblock Auto-encoding variational {Bayes}.
\newblock In \emph{International Conference on Learning Representations}, 2014.

\bibitem[Kurakin et~al.(2016)Kurakin, Goodfellow, and
  Bengio]{kurakin2016adversarial}
Kurakin, A., Goodfellow, I., and Bengio, S.
\newblock Adversarial machine learning at scale.
\newblock \emph{arXiv preprint arXiv:1611.01236}, 2016.

\bibitem[Li et~al.(2019)Li, Bradshaw, and Sharma]{li2018generative}
Li, Y., Bradshaw, J., and Sharma, Y.
\newblock Are generative classifiers more robust to adversarial attacks?
\newblock In \emph{International Conference on Machine Learning}, 2019.

\bibitem[Madry et~al.(2017)Madry, Makelov, Schmidt, Tsipras, and
  Vladu]{madry2017towards}
Madry, A., Makelov, A., Schmidt, L., Tsipras, D., and Vladu, A.
\newblock Towards deep learning models resistant to adversarial attacks.
\newblock \emph{International Conference on Learning Representations}, 2017.

\bibitem[Pang et~al.(2018)Pang, Du, and Zhu]{pang2018max}
Pang, T., Du, C., and Zhu, J.
\newblock Max-mahalanobis linear discriminant analysis networks.
\newblock In \emph{International Conference on Machine Learning}, pp.\
  4016--4025, 2018.

\bibitem[Pang et~al.(2019)Pang, Xu, Du, Chen, and Zhu]{pang2019improving}
Pang, T., Xu, K., Du, C., Chen, N., and Zhu, J.
\newblock Improving adversarial robustness via promoting ensemble diversity.
\newblock In \emph{International Conference on Machine Learning}, 2019.

\bibitem[Pang et~al.(2020{\natexlab{a}})Pang, Xu, Dong, Du, Chen, and
  Zhu]{pang2020rethinking}
Pang, T., Xu, K., Dong, Y., Du, C., Chen, N., and Zhu, J.
\newblock Rethinking softmax cross-entropy loss for adversarial robustness.
\newblock In \emph{International Conference on Learning Representations},
  2020{\natexlab{a}}.

\bibitem[Pang et~al.(2020{\natexlab{b}})Pang, Xu, and Zhu]{pang2020mixup}
Pang, T., Xu, K., and Zhu, J.
\newblock Mixup inference: Better exploiting mixup to defend adversarial
  attacks.
\newblock In \emph{International Conference on Learning Representations},
  2020{\natexlab{b}}.

\bibitem[Papernot et~al.(2016{\natexlab{a}})Papernot, Faghri, Carlini,
  Goodfellow, Feinman, Kurakin, Xie, Sharma, Brown, Roy,
  et~al.]{papernot2016technical}
Papernot, N., Faghri, F., Carlini, N., Goodfellow, I., Feinman, R., Kurakin,
  A., Xie, C., Sharma, Y., Brown, T., Roy, A., et~al.
\newblock Technical report on the {CleverHans} {v2.1.0} adversarial examples
  library.
\newblock \emph{arXiv preprint arXiv:1610.00768}, 2016{\natexlab{a}}.

\bibitem[Papernot et~al.(2016{\natexlab{b}})Papernot, McDaniel, and
  Goodfellow]{papernot2016transferability}
Papernot, N., McDaniel, P., and Goodfellow, I.
\newblock Transferability in machine learning: from phenomena to black-box
  attacks using adversarial samples.
\newblock \emph{arXiv preprint arXiv:1605.07277}, 2016{\natexlab{b}}.

\bibitem[Papernot et~al.(2016{\natexlab{c}})Papernot, McDaniel, Wu, Jha, and
  Swami]{papernot2016distillation}
Papernot, N., McDaniel, P., Wu, X., Jha, S., and Swami, A.
\newblock Distillation as a defense to adversarial perturbations against deep
  neural networks.
\newblock In \emph{2016 IEEE Symposium on Security and Privacy (SP)}, pp.\
  582--597. IEEE, 2016{\natexlab{c}}.

\bibitem[Prakash et~al.(2018)Prakash, Moran, Garber, DiLillo, and
  Storer]{prakash2018deflecting}
Prakash, A., Moran, N., Garber, S., DiLillo, A., and Storer, J.
\newblock Deflecting adversarial attacks with pixel deflection.
\newblock In \emph{Proceedings of the IEEE conference on computer vision and
  pattern recognition}, pp.\  8571--8580, 2018.

\bibitem[Rauber et~al.(2017)Rauber, Brendel, and Bethge]{rauber2017foolbox}
Rauber, J., Brendel, W., and Bethge, M.
\newblock Foolbox: A python toolbox to benchmark the robustness of machine
  learning models.
\newblock \emph{arXiv preprint arXiv:1707.04131}, 2017.

\bibitem[Roth et~al.(2019)Roth, Kilcher, and Hofmann]{roth2019odds}
Roth, K., Kilcher, Y., and Hofmann, T.
\newblock The odds are odd: A statistical test for detecting adversarial
  examples.
\newblock In \emph{International Conference on Machine Learning}, 2019.

\bibitem[Sabour et~al.(2016)Sabour, Cao, Faghri, and
  Fleet]{sabour2015adversarial}
Sabour, S., Cao, Y., Faghri, F., and Fleet, D.~J.
\newblock Adversarial manipulation of deep representations.
\newblock \emph{International Conference on Learning Representations}, 2016.

\bibitem[Santurkar et~al.(2019)Santurkar, Tsipras, Tran, Ilyas, Engstrom, and
  Madry]{santurkar2019computer}
Santurkar, S., Tsipras, D., Tran, B., Ilyas, A., Engstrom, L., and Madry, A.
\newblock Computer vision with a single (robust) classifier.
\newblock In \emph{Advances in Neural Information Processing Systems}, 2019.

\bibitem[Schott et~al.(2019{\natexlab{a}})Schott, Rauber, Bethge, and
  Brendel]{schott2018towards}
Schott, L., Rauber, J., Bethge, M., and Brendel, W.
\newblock Towards the first adversarially robust neural network model on
  {MNIST}.
\newblock In \emph{International Conference on Learning Representations},
  2019{\natexlab{a}}.

\bibitem[Schott et~al.(2019{\natexlab{b}})Schott, Rauber, Brendel, and
  Bethge]{lukas_abs}
Schott, L., Rauber, J., Brendel, W., and Bethge, M.
\newblock Robust perception through analysis by synthesis.
\newblock \emph{International Conference on Learning Representations},
  2019{\natexlab{b}}.

\bibitem[Sen et~al.(2020)Sen, Ravindran, and Raghunathan]{sen2020empir}
Sen, S., Ravindran, B., and Raghunathan, A.
\newblock {\{}EMPIR{\}}: Ensembles of mixed precision deep networks for
  increased robustness against adversarial attacks.
\newblock In \emph{International Conference on Learning Representations}, 2020.

\bibitem[Szegedy et~al.(2014)Szegedy, Zaremba, Sutskever, Bruna, Erhan,
  Goodfellow, and Fergus]{szegedy2013intriguing}
Szegedy, C., Zaremba, W., Sutskever, I., Bruna, J., Erhan, D., Goodfellow, I.,
  and Fergus, R.
\newblock Intriguing properties of neural networks.
\newblock In \emph{International Conference on Learning Representations
  (ICLR)}, 2014.

\bibitem[Tram{\`e}r \& Boneh(2019)Tram{\`e}r and Boneh]{tramer2019adversarial}
Tram{\`e}r, F. and Boneh, D.
\newblock Adversarial training and robustness for multiple perturbations.
\newblock In \emph{Advances in Neural Information Processing Systems}, pp.\
  5858--5868, 2019.

\bibitem[Tram\`er et~al.(2018)Tram\`er, Kurakin, Papernot, Goodfellow, Boneh,
  and McDaniel]{tramer2017ensemble}
Tram\`er, F., Kurakin, A., Papernot, N., Goodfellow, I., Boneh, D., and
  McDaniel, P.
\newblock Ensemble adversarial training: Attacks and defenses.
\newblock In \emph{International Conference on Learning Representations}, 2018.

\bibitem[Tsipras et~al.(2019)Tsipras, Santurkar, Engstrom, Turner, and
  Madry]{tsipras2018robustness}
Tsipras, D., Santurkar, S., Engstrom, L., Turner, A., and Madry, A.
\newblock Robustness may be at odds with accuracy.
\newblock In \emph{International Conference on Learning Representations}, 2019.

\bibitem[Uesato et~al.(2018)Uesato, O'Donoghue, Oord, and
  Kohli]{uesato2018adversarial}
Uesato, J., O'Donoghue, B., Oord, A. v.~d., and Kohli, P.
\newblock Adversarial risk and the dangers of evaluating against weak attacks.
\newblock \emph{arXiv preprint arXiv:1802.05666}, 2018.

\bibitem[Verma \& Swami(2019)Verma and Swami]{verma2019error}
Verma, G. and Swami, A.
\newblock Error correcting output codes improve probability estimation and
  adversarial robustness of deep neural networks.
\newblock In \emph{Advances in Neural Information Processing Systems}, pp.\
  8643--8653, 2019.

\bibitem[Willetts et~al.(2019)Willetts, Camuto, Roberts, and
  Holmes]{willetts2019disentangling}
Willetts, M., Camuto, A., Roberts, S., and Holmes, C.
\newblock Disentangling improves vaes' robustness to adversarial attacks, 2019.

\bibitem[Xiao et~al.(2020)Xiao, Zhong, and Zheng]{xiao2019resisting}
Xiao, C., Zhong, P., and Zheng, C.
\newblock Resisting adversarial attacks by $k$-winners-take-all.
\newblock In \emph{International Conference on Learning Representations}, 2020.

\bibitem[Yang et~al.(2019{\natexlab{a}})Yang, Zhang, Katabi, and
  Xu]{yang2019me}
Yang, Y., Zhang, G., Katabi, D., and Xu, Z.
\newblock Me-net: Towards effective adversarial robustness with matrix
  estimation.
\newblock In \emph{International Conference on Machine Learning}, pp.\
  7025--7034, 2019{\natexlab{a}}.

\bibitem[Yang et~al.(2019{\natexlab{b}})Yang, Li, Chen, and
  Song]{yang2018characterizing}
Yang, Z., Li, B., Chen, P.-Y., and Song, D.
\newblock Characterizing audio adversarial examples using temporal dependency.
\newblock In \emph{International Conference on Learning Representations},
  2019{\natexlab{b}}.

\bibitem[Yin et~al.(2020)Yin, Kolouri, and Rohde]{yin2020adversarial}
Yin, X., Kolouri, S., and Rohde, G.~K.
\newblock Adversarial example detection and classification with asymmetrical
  adversarial training.
\newblock In \emph{International Conference on Learning Representations}, 2020.

\end{thebibliography}
\bibliographystyle{icml2020}

\end{document}